\documentclass[journal]{IEEEtran}
\usepackage{amsmath,amsfonts}
\usepackage{algorithmic}
\usepackage{algorithm}
\usepackage{array}
\usepackage[caption=false,font=normalsize,labelfont=sf,textfont=sf]{subfig}
\usepackage{textcomp}
\usepackage{stfloats}
\usepackage{url}
\usepackage{verbatim}
\usepackage{graphicx}
\usepackage{cite}
\usepackage{romannum}
\usepackage{tabularx}
\usepackage{multirow}
\usepackage{changepage} 
\usepackage{threeparttable} 
\usepackage{pgfplots}
\usepackage{orcidlink} 
\usepackage{tikz}
\newcommand\submittedtext{%
  \footnotesize This work has been submitted to the IEEE Transactions on Affective Computing for possible publication. Copyright may be transferred without notice, after which this version may no longer be accessible.}

\newcommand\submittednotice{%
\begin{tikzpicture}[remember picture,overlay]
\node[anchor=south,yshift=10pt] at (current page.south) {\fbox{\parbox{\dimexpr0.65\textwidth-\fboxsep-\fboxrule\relax}{\submittedtext}}};
\end{tikzpicture}%
}
\pgfplotsset{compat=1.18}

\begin{document}

 \title{AMuSeD: An Attentive Deep Neural Network for Multimodal Sarcasm Detection Incorporating Bi-modal Data Augmentation}

\author{Xiyuan Gao\,\orcidlink{0000-0003-0870-6721}, Shubhi Bansal\,\orcidlink{0000-0002-8034-8220}, Kushaan Gowda\,\orcidlink{0000-0001-6921-2069}, Zhu Li\, \orcidlink{0000-0002-1409-2482}, Shekhar Nayak\,\orcidlink{0000-0002-4277-4851}, Nagendra Kumar\,\orcidlink{0000-0003-4644-3168}, Matt Coler\,\orcidlink{0000-0002-7631-5063}

\thanks{Manuscript submitted on June 13, 2024. \textit{(Corresponding author: Xiyuan Gao)}}
\thanks{X.Gao, Z.Li, S.Nayak and M.Coler are with Campus Fryslân, University of Groningen, Leeuwarden 8911 CE, the Netherlands (e-mails: xiyuan.gao@rug.nl; zhu.li@rug.nl; s.nayak@rug.nl; m.coler@rug.nl)}
\thanks{S.Bansal and N.Kumar are with Computer Science and Engineering, Indian Institute of Technology Indore, Indore 453552, India (e-mail: phd2001201007@iiti.ac.in; nagendra@iiti.ac.in).}
\thanks{K.Gowda was with Computer Science and Engineering, Indian Institute of Technology Indore, Indore 453552, India. He is now with Computer Science, Columbia University, New York 10027, USA (e-mail: kg3081@columbia.edu)}}
\maketitle
\submittednotice
\thispagestyle{empty}

\begin{abstract}
Detecting sarcasm effectively requires a nuanced understanding of context, including vocal tones and facial expressions. The progression towards multimodal computational methods in sarcasm detection, however, faces challenges due to the scarcity of data. To address this, we present AMuSeD (Attentive deep neural network for MUltimodal Sarcasm dEtection incorporating bi-modal Data augmentation). This approach utilizes the Multimodal Sarcasm Detection Dataset (MUStARD) and introduces a two-phase bimodal data augmentation strategy. The first phase involves generating varied text samples through Back Translation from several secondary languages. The second phase involves the refinement of a FastSpeech 2-based speech synthesis system, tailored specifically for sarcasm to retain sarcastic intonations. Alongside a cloud-based Text-to-Speech (TTS) service, this Fine-tuned FastSpeech 2 system produces corresponding audio for the text augmentations. We also investigate various attention mechanisms for effectively merging text and audio data, finding self-attention to be the most efficient for bimodal integration. Our experiments reveal that this combined augmentation and attention approach achieves a significant F1-score of 81.0\% in text-audio modalities, surpassing even models that use three modalities from the MUStARD dataset. 
\end{abstract}


\section{Introduction}

\begin{figure}
	\centering
\includegraphics[width=0.5\textwidth]{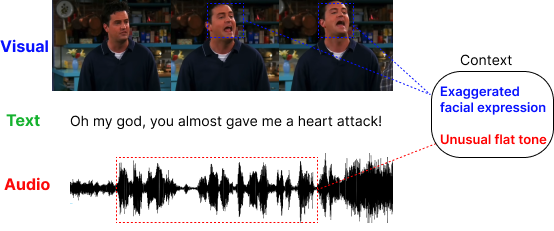}
\caption{Sample sarcastic utterance from MUStARD dataset.}
	\label{chandler}
\end{figure}

\IEEEPARstart{S}{arcasm} is a complex linguistic phenomenon that is not straightforward to interpret. It frequently communicates feelings that are the opposite of what the words literally mean, using a combination of verbal and non-verbal signals. This complexity poses significant challenges for computational detection, requiring an approach that integrates multiple modes of information. Sarcasm’s multidisciplinary nature spans various fields, including linguistics and psychology. Research in these fields has identified cues that are essential for understanding of sarcasm, including tone of voices or facial expressions~\cite{Lishapeng}, exaggeration of statements~\cite{kreuz1}, expectations and the common ground between speakers and hearers~\cite{kreuz2}. Drawing insights from these disciplines is crucial for the computational analysis of sarcasm. To better understand sarcasm’s nuanced attributes, we present an example from the TV sitcom ‘Friends’, as depicted in Figure~\ref{chandler}. In this case, the character Chandler feigns surprise in such a way that the listener knows that he is being insincere. We highlight two key aspects of sarcasm: first, there is a divergence between the speaker's actual intention and the literal meaning of their words; second, as Sperber and Wilson~\cite{Wilson} pointed out, sarcasm serves as a tool to communicate underlying opinions. This leads us to another aspect of sarcasm: the importance of context in deciphering the speaker’s intent. As illustrated in Figure~\ref{chandler}, without context, in this case, an unusual flat tone and exaggerated facial expression, Chandler’s comment could be misconstrued as genuine surprise upon seeing Joey emerge from the box. However, the sarcastic nature of his statement becomes clear with his tone and facial expression. Context in sarcasm often involves explicit cues like variations in tone, exaggerated emphasis, extended syllables, or a serious facial expression. It can also be implicit, drawing on shared cultural understanding~\cite{Gibbs, Kreuz}.

Sarcasm represents a significant challenge particularly in fields like Sentiment Analysis and Speech Emotion Recognition (SER). The rise of rich multimedia communication platforms has highlighted the shortcomings of unimodal analyses, thereby shifting the focus towards multimodal methods that integrate text, visual, and audio data. A unimodal approach can erroneously classify utterances, such as the one illustrated in Figure~\ref{chandler}, as “surprise” or “positive” without taking into account contextual cues, such as tone of voice or facial expressions, derived from the audio and visual modalities. Multimodal sarcasm detection, such as that of Schifanella~\textit{et al}.~\cite{Schifanella} delved into the complex interplay between text and visual data in multimodal contexts, emphasizing the need for a multimodal strategy. An important development in this area was the creation of the MUStARD dataset by Castro~\textit{et al}.~\cite{Castro}, as it is a primary multimodal dataset containing sarcasm in conversations. This dataset, featuring text, audio, and video data from American TV sitcoms, proved the viability and effectiveness of multimodal data in sarcasm detection. Subsequent research has concentrated on refining methods for integrating these different modalities and improving fusion techniques to enhance detection accuracy. For instance, Wu~\textit{et al}.~\cite{Wu} introduced an incongruity-aware attention network that assesses discordance across modalities, prioritizing segments with higher incongruity. This approach, when applied to the MUStARD dataset, significantly improved sarcasm detection. Similarly, Pramanick~\textit{et al}.~\cite{Pramanick} used the same dataset to implement a self-attention mechanism~\cite{Vaswani} that captures both intra-modal and cross-modal relationships.

Despite the progress in multimodal research, the challenge of data scarcity remains a significant barrier. A notable instance of this is the MUStARD dataset, which only includes about 690 utterances, spanning multiple modalities, and with an average audio clip duration of merely 5.2 seconds, cumulatively amounting to less than one hour of speech data. Our approach aims to mitigate this limitation by introducing bimodal data augmentation techniques and enhancing feature fusion through attention mechanisms. Data augmentation is widely recognized for increasing model robustness and preventing overfitting. While it has been extensively applied to unimodal data~\cite{Rico, Marivate, Ko, Li}, its application in bimodal data augmentation remains relatively unexplored~\cite{Xu, Ruixue}. To our knowledge, the intersection of text and audio in the context of bimodal data augmentation is yet to be fully explored. Moreover, the use of self-attention mechanisms in processing extracted features has been shown to be an effective strategy in recent studies~\cite{Pramanick, Kumar, Ray}. These mechanisms are instrumental in identifying and enhancing features that are crucial for sarcasm detection in each modality during the training process, thereby enhancing overall model performance. Building on these foundations, our research proposes the following research questions (RQs).

\textbf{RQ1:} Does our innovative text-audio data augmentation approach enhance the detection of sarcasm in a multimodal context, and which key factors primarily determine its success?

\textbf{RQ2:} How does incorporating self-attention mechanisms with our augmentation strategy refine the model’s performance, thus pushing the boundaries of multimodal sarcasm detection?

To address RQ1, we implement a novel data augmentation technique, using Back Translation to expand our text corpus, and speech synthesis to augment our audio data. We focus on two critical aspects: the quantity of augmented data and the effectiveness of fine-tuned speech synthesizers compared to standard models. Addressing RQ2 involves evaluating the influence of the self-attention mechanism on our augmented dataset, particularly its ability to selectively enhance features relevant to sarcasm detection. The main contributions of this research work are:
\begin{itemize}
\item{We develop a text-audio bimodal data augmentation technique that significantly enhances sarcasm detection.}
\item{We validate the effectiveness of self-attention mechanisms in improving feature extraction for multimodal sarcasm detection.}
\item{We advance FastSpeech 2’s capabilities for nuanced sarcasm expression, enriching audio quality in augmented datasets.}
\item{We introduce AMuSeD, a groundbreaking framework for multimodal sarcasm detection, achieving an F1-score of 81.0\% on the MUStARD dataset.}
\end{itemize}

The structure of the paper is as follows: Section \Romannum{2} provides an overview of the relevant literature, focusing on multimodal sarcasm detection and bimodal data augmentation. Section \Romannum{3} details our proposed methodology, including the augmentation process, feature extraction, and fusion. Section \Romannum{4} describes the experimental setup and presents the findings. Section \Romannum{5} discusses these results. The paper concludes with Section \Romannum{6}.

\section{Related Work}
We categorize this review into two parts. The first explores multimodal sarcasm detection, examining the integration of textual and auditory data. This area presents unique challenges and opportunities for methodological innovation. In the second part, we assess the efficacy of our approach by exploring advanced text and audio data augmentation techniques and their impact on each modality.

\subsection{Multimodal Sarcasm Detection}
In multimodal sarcasm detection, we find that integrating diverse communication modes (text, audio, visual cues) yields a more nuanced understanding of sarcasm. The integration of multiple modalities has been a major focus of research in multimodal sarcasm detection. In this section, we discuss multimodal sarcasm detection based on three major fusion approaches: a late fusion approach, an early fusion approach, and attention-based fusion. 

\subsubsection{Late Fusion Approach} In late fusion, text and audio features are processed independently and then combined for final classification. Schifanella~\textit{et al}.~\cite{Schifanella} employed this method in their study on the interplay of text and images on social media. They noted the crucial role of these modalities in conveying contrasting messages for sarcasm detection. Utilizing MUStARD dataset, Ding~\textit{et al}.~\cite{Ding} proposed a fusion framework that concatenated the post-processed features from three modalities (i.e., text, audio, video). For each of the modality, features were extracted and went through a fully-connected neural network called SubNet. Late fusion approach offers flexibility, allowing each modality to utilize optimal models for feature extraction and classification. However, it may not fully capture the interactions between low-level features across different modalities.

\subsubsection{Early Fusion Approach} This method involves merging features from various modalities prior to model training. For instance, Castro~\textit{et al}.~\cite{Castro} combined text, audio, and video features, feeding them into a Support Vector Machine (SVM) to assess the efficacy of multimodal data integration in sarcasm detection. Hiremath and Patil~\cite{Hiremath} concatenated text, audio, and video features and sent them to a fully-connected neural network to classify sarcasm and non-sarcasm utterances. This approach enables the model to identify correlations between low-level features. Yet, integrating disparate features effectively remains a challenge. 

\subsubsection{Attention-based Fusion} Increasingly popular in multimodal sarcasm detection, attention-based fusion focuses on the relationships within and between modalities using attention mechanisms. It assigns greater weights to more closely related features, potentially enhancing classification accuracy. For example, Chauhan~\textit{et al}.~\cite{Chauhan} applied attention mechanisms to learn the relationship between the feature vector of a segment of an utterance in one modality and another segment of the utterance in a different modality. Zhang~\textit{et al}.~\cite{XZhang} utilized contrastive attention mechanism to capture the irrelevance between two feature vectors from different modalities. In addition, Zhang~\textit{et al}.~\cite{YZhang} employed a self attention mechanism to capture the important contextual information of the targeting utterance in each modality, and further built a cross-modal attention mechanism among three modalities to enhance the sarcasm-related features. Attention-based fusion offers nuanced feature weighting but requires careful calibration to ensure efficacy. 

We contend that the complexity of sarcasm transcends textual analysis, and requires an intricate fusion of multiple communicative cues. The challenge lies not only in capturing the essence of sarcasm from each individual modality – be it text, audio, or visual cues – but also in effectively synthesizing these disparate elements. Exploring the integration of these modalities offers insights into the potential of multimodal analysis in overcoming the limitations of unimodal methods, paving the way for more sophisticated and accurate sarcasm detection systems. 

\subsection{Data Augmentation in Multimodal Sarcasm Detection}
Data augmentation, a process of generating additional samples by transforming existing training data, plays a crucial role in enhancing machine learning models’ generalization capabilities and robustness. Its significance spans various domains, including image processing, computer vision, audio, and text processing. This subsection considers textual data augmentation, audio data augmentation, and multimodal data augmentation, respectively.

\subsubsection{Textual Data Augmentation} 
In text processing, diverse techniques are employed for specific objectives. For instance, Back Translation, popularized by Rico~\textit{et al}.~\cite{Rico}, involves translating text to a different language and then back to the original. This method maintains the linguistic essence while altering word choice and syntax, proving effective in maintaining label consistency and paraphrasing. Aroyehun and Gelbukh~\cite{Aroyehun} and Marivate and Sefara~\cite{Marivate} used this technique with languages like French, Spanish, and German to enhance datasets in aggression and hate speech detection. Lee~\textit{et al}.~\cite{Lee} applied it in sarcasm detection to generate non-sarcastic samples, showcasing its versatility across various natural language processing (NLP) applications.

\subsubsection{Audio Data Augmentation}
In sarcasm detection, Gao~\textit{et al}.~\cite{Gao} demonstrated the effectiveness of speed and volume perturbations. While sarcasm-specific research in audio augmentation is limited, related areas like Automatic Speech Recognition and SER have extensively used methods like time shift, pitch shift, and random noise injection. Advances in speech synthesis, such as Generative Adversarial Networks and Tacotron2, offer novel augmentation possibilities. Li~\textit{et al}.~\cite{Li} initiated this approach, employing Tacotron 2 as the TTS model to capture speaker identities from the multi-speaker LibriTTS dataset~\cite{LibriTTS}. Subsequently, Rosenberg~\textit{et al}.~\cite{Rosenberg} uses a Tacotron 2-based multispeaker model upon the dataset LibriTTS in different domains, to demonstrate the validity of applying this augmentation method in domain transfer.   Notably, synthesized speech still shows performance gaps compared to human speech in recognition tasks, but exhibits promise in replicating specific speech characteristics like sarcasm.

\subsubsection{Multimodal Data Augmentation}
In multimodal sarcasm detection, challenges persist due to data scarcity. Huang~\textit{et al}.~\cite{Huang} introduced a method for augmenting data in multimodal emotion classification by segmenting original sequences, though this may overlook inherent sequence associations. In the text-image domain, Xu~\textit{et al}.~\cite{Xu} and Ruixue~\textit{et al}.~\cite{Ruixue} employed generative networks for creating text-image pairs and augmenting image data, demonstrating the significance of semantic consistency in applications like Visual Question Answering. Hao~\textit{et al}.~\cite{Hao} introduced MixGen, which generates augmented samples by combining images and text. However, text-audio augmentation specifically for sarcasm detection remains unexplored.

While attention mechanisms in modality fusion reveal intricate inter-modal relationships, data augmentation techniques like Back Translation and synthesized speech show potential in enhancing model performance. The synergistic application of these methods in text-audio bimodal data augmentation, particularly in sarcasm detection, holds promise for future research.

Transitioning from the foundational understanding of multimodal sarcasm detection, and the challenges and innovations highlighted in bimodal data augmentation and feature fusion, next we present our methodology. This methodology encapsulates the strategic integration of text and audio modalities, employing advanced techniques in data augmentation and feature extraction. Our approach, Attentive Deep Neural Network for MUltimodal Sarcasm dEtection incorporating Bi-modal Data Augmentation (AMuSeD), leverages the strengths of both text and audio data to capture the multifaceted nature of sarcasm more effectively. 

\begin{figure*}
	\centering
\includegraphics[width=\textwidth]{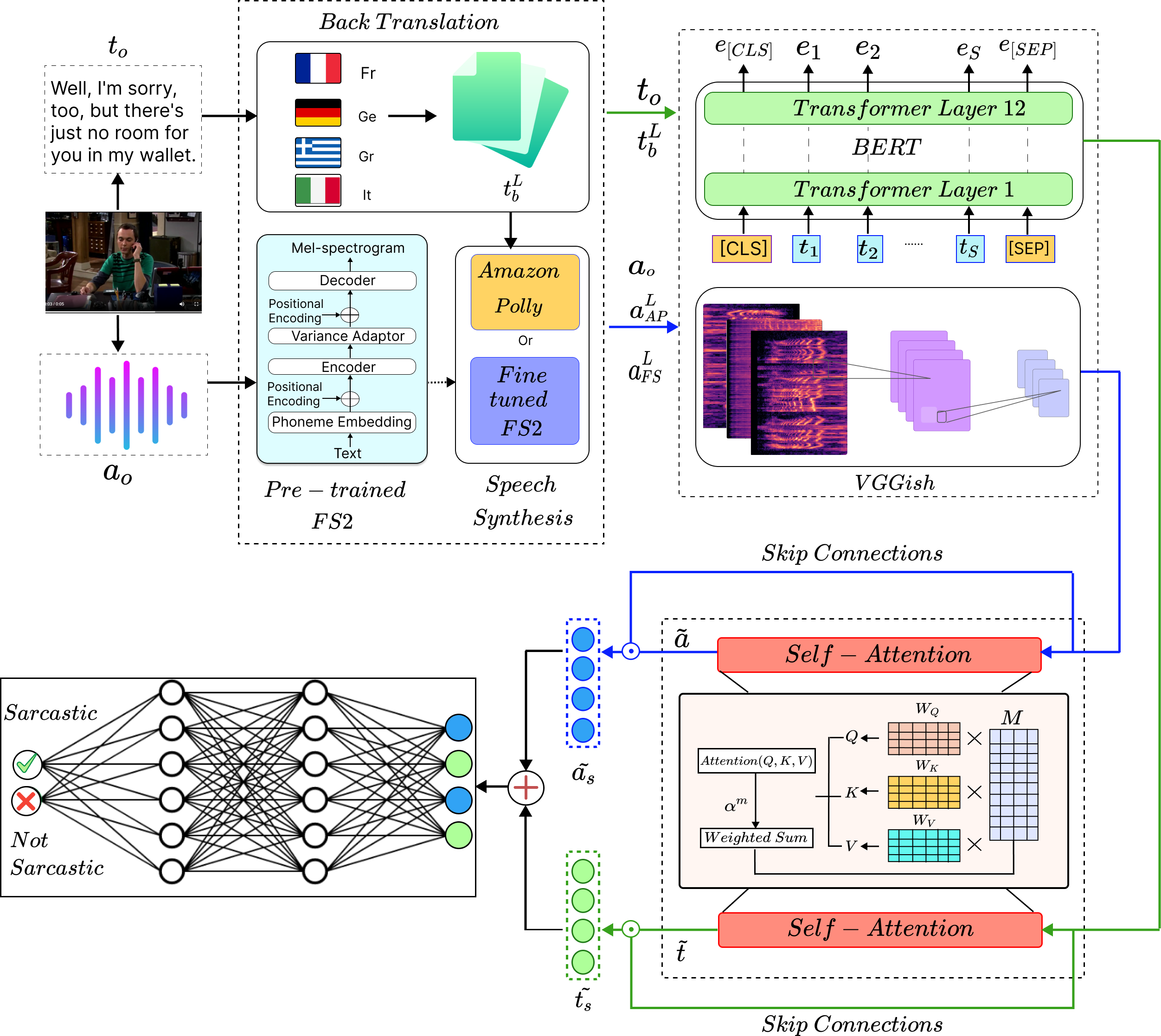}
\caption{System architecture of the proposed AMuSeD.}
	\label{fig:arch}
\end{figure*}

\section{Methodology}
We introduce an Attentive deep neural network for MUltimodal Sarcasm dEtection incorporating bi-modal Data augmentation (AMuSeD), a sarcasm detection methodology that integrates text and audio modalities. The architecture of AMuSeD is represented in Figure~\ref{fig:arch}, which provides an overview of the system architecture and components. This approach comprises the following steps:

\begin{enumerate}
\item{Data augmentation: We commence by generating augmented data for both text and audio through offline techniques. For text, we utilize Back Translation as described by Rico~\textit{et al}.~\cite{Rico}. For audio, we employ speech synthesis to augment the dataset.}
\item{Feature extraction: Leveraging state-of-the-art pre-trained models, we extract features from each modality. BERT~\cite{bert} is for text to derive high-dimensional embeddings capturing linguistic nuances. For audio, we utilize VGGish~\cite{Hershey}, a model adept at extracting salient audio embeddings.}
\item{Feature fusion: The core of our methodology lies in the fusion of extracted features. We employ a self-attention mechanism, augmented with skip connections, to synergize the text and audio embeddings. This step allows the model to focus on relevant features from both modalities, enhancing its ability to detect sarcasm.}
\item{Prediction: The fused features are then concatenated and passed through a fully-connected layer. This layer serves as the decision-making component, outputting the final sarcasm label.}
\end{enumerate}

This remainder of this section is structured as follows: Subsection $A$ describes our data augmentation strategy. Subsection $B$ explains the feature extraction method. In addition, subsection $C$ illustrates the mechanisms of fusion and classification. 

\subsection{Data Augmentation Strategy}
\begin{figure*}
	\centering
\includegraphics[width=0.8\textwidth]{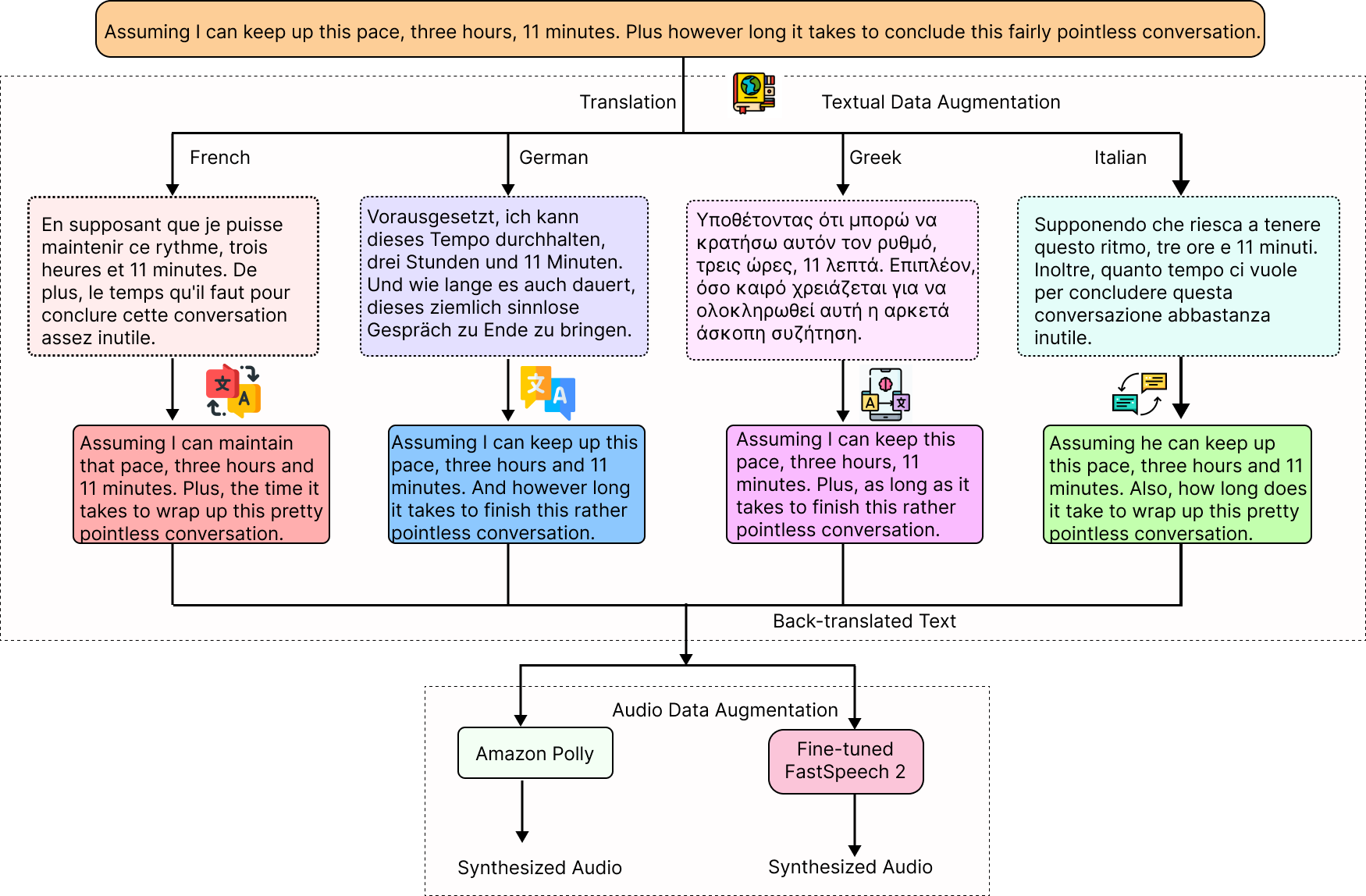}
	\caption{\centering Schematic overview of data augmentation.}
	\label{fig:da}
\end{figure*}

Our data augmentation approach addresses data scarcity challenges in sarcasm detection by synchronizing text and audio modalities to maintain semantic coherence, as depicted in Figure~\ref{fig:da}. This strategy involves the following three steps.

\subsubsection{Step 1: Text Augmentation with Back Translation}

This process involves generating paraphrased versions of the original sarcastic statements by translating them into a different language and then re-translating them back to the original language. The rationale behind using Back Translation lies in its effectiveness in preserving the sarcasm inherent in the text. Our initial training dataset consists of 690 samples. The back-translation process itself is bifurcated into two primary phases. The first phase involves translating the original text  $t_o$ into a secondary language. To those ends, we employ back-translation, as described below:
\begin{equation}
\label{deqn_ex1}
t_b^L=Back Translation(t_o)
\end{equation}
Here, $t_o$ denotes the original English language sample (source language) that is translated into a set of chosen target languages, represented by $L$. The notation $t_b^L$ signifies the samples that have been back-translated from these target languages $L$ to English. In our study, $L$ encompasses Greek ($Gr$), German ($Ge$), French ($Fr$), and Italian ($Ita$) as the target languages. These languages were selected because languages closely resembling English tend to yield numerous duplicates post back-translation, diminishing the effectiveness of the experiment. Conversely, languages very remote from English could introduce significant translation errors. Thus, the chosen languages are an attempt to balance minimizing duplicates and translation inaccuracies.

Once the translation is completed, we reverse the process, translating the data back to English. We utilize Googletrans\footnote{https://pypi.org/project/googletrans/}, a tool that leverages the Google Translate Ajax API, for translating between the source and target languages. Googletrans employs Google’s neural machine translation, which uses a transformer-based encoder and a Recurrent Neural Network decoder. Our initial dataset comprised 690 samples. Post back-translation, we obtained 690 samples for each secondary language, including duplicates that emerged during the process. To ensure model robustness and prevent bias, we systematically removed these duplicates across languages by iteratively comparing utterance pairs. See Table~\ref{table1}. 

\begin{table}
\caption{Back-translated Text\label{table1}}
\centering
\small  
\begin{tabularx}{0.45\textwidth}{|X|X|}
\hline
\textbf{Secondary Languages} & \textbf{Sample Numbers} \\ [0.5ex] 
\hline\hline
Greek & 544 \\
\hline
German & 596 \\
\hline
French & 476 \\
\hline
Italian & 476 \\
\hline
\end{tabularx}
\end{table}

\subsubsection{Step 2: Speech Augmentation}
Our approach to speech-based data augmentation uses speech synthesis and involves generating natural-sounding speech from text. Given the scarcity of sarcastic speech data for training a TTS system from scratch, we explore two alternative speech synthesis models: Amazon Polly and FastSpeech 2.

\textbf {Amazon Polly (AP)\footnote{https://docs.aws.amazon.com/polly/latest/dg/what-is.html}}: AP is a neural TTS service by Amazon Web Services. It converts text into lifelike speech across various languages and voices using advanced deep learning techniques. Although the exact size and composition of AP’s training dataset are undisclosed, it is known to be extensive, incorporating multilingual and multi-speaker data to achieve natural-sounding speech synthesis. AP offers a range of languages, male and female voices, accents, and tones. For our study, we selected four voice identities: Emma (English-GB), Sally (English-US), Brian (English-GB), and Joey (English-US), with the first two being female and the latter two male voices.

\textbf{FastSpeech 2 (FS2)}: FS2 proposed by Chien~\textit{et al}.~\cite{FastSpeech}, is a sequence-to-sequence TTS model that predicts mel spectrogram frames directly, facilitating parallelization and faster synthesis. Its architecture includes an encoder for text encoding, a duration predictor for phoneme timing, and a decoder for generating the mel spectrogram frames. We first pre-trained FS2 on a diverse dataset, such as LibriTTS, to generate high-quality speech. We then fine-tuned it on the MUStARD dataset to enhance its capability in conveying sarcastic features.

With the following content, we provide the implementation details of FS2:

\begin{itemize}
    \item{
Pre-training FS2: We used a Pre-trained FS2 model to generate speech for sarcasm detection, initially pre-training it on the LibriTTS dataset, which contains recordings from multiple speakers. The synthesized speech was based on back-translated text.}

\item{Fine-tuning FS2: This process involved refining the Pre-trained FS2 model using the MUStARD dataset. We resampled all speech to 22.05kHz and extracted 80-dimensional mel-spectrograms with a filter length of 1,024 and hop length of 256. Phoneme durations were obtained using the Montreal Forced Aligner tool\footnote{https://montreal-forced-aligner.readthedocs.io/en/latest/}. Employing an open-source FS2 model\footnote{https://github.com/ming024/FastSpeech2}, we conducted 800,000 iterations for pre-training and 200,000 for fine-tuning. The training used the Adam~\cite{Kingma} optimizer with specific parameters $\beta_1=0.9$, $\beta_2=0.98$, $\epsilon=10^{-9}$, and a warm-up strategy was implemented for the first 4,000 iterations. Additionally, a well-trained HiFi-GAN~\cite{HiFi} was utilized as the vocoder.}
\end{itemize}

\subsubsection{Step 3: Text-audio Biomodal Data Augmentation}
Our approach to text-audio bimodal data augmentation involves an initial phase of text back-translation in various secondary languages $L=\{Gr, Ge, Fr, Ita\}$, resulting in augmented texts $t_b^L$. Subsequently, corresponding audio samples $a^L$ are generated using Amazon Polly (AP) and FastSpeech 2 (FS2).

\begin{subequations}\label{eq:2}
\begin{align}
t_b^L=Back Translation(t_o)\label{eq:2A}\\
a_{AP}^L=Amazon Polly(t_b^L)\label{eq:2B}\\
a_{FS}^L=FastSpeech 2(t_b^L)\label{eq:2c}
\end{align}
\end{subequations}

To investigate RQ1, which pertains to the factors influencing the efficacy of our data augmentation method, we create augmented audio datasets of various sizes to examine the impact of data volume. These datasets encompass 4-fold, 16-fold, and 20-fold augmentation. It is critical to align all audio samples with their corresponding text during augmentation. For computational efficiency, we restrict the comparison of the three selected synthesis methods (AP, Pre-trained FS2, Fine-tuned FS2) to the 4-fold augmented dataset.

\textbf{4-fold Augmentation:} This dataset aims to compare the effectiveness of different speech synthesizers, comprising three types of audio samples:

\begin{itemize}
    \item{2,528 samples generated by applying four AP speaker IDs (Brian, Emma, Joey, Salli) to 632 back-translated texts in Latin\footnote{Latin is used only in preliminary experiments and has since been removed from our inventory due to its high similarity to other secondary languages, which could result in a large number of duplicates.}}
    \item{2,092 samples produced using the Pre-trained FS2, derived from back-translated texts in Greek, German, French, and Italian.}
    \item{An equivalent set of 2,092 samples created with the Fine-tuned FS2 model, using the same secondary languages.}
\end{itemize}

\textbf{16-fold Augmentation:} This dataset, comprising 8,368 samples, is generated using the same four AP speaker IDs (Brian, Emma, Joey, Salli), based on 2,092 back-translated texts in Greek, German, French, and Italian.

\textbf{20-fold Augmentation:} The largest dataset, with 10,460 samples, is formed by combining all audio from the 4-fold Fine-tuned FS2 samples and the 16-fold AP samples. This dataset aims to test the compatibility of audios synthesized by different models.

It is important to note that the terminology used (e.g., 4-fold) does not directly reflect the literal sample count, due to the removal of duplicates in the text modality. To ensure clarity and transparency, we have described the terminology and corresponding sample numbers in Table~\ref{table2}. 

\begin{table}[H]
\caption{Augmented audio samples\label{table2}}
\centering
\small
\begin{tabular}{|m{8em} | m{16em}|}
\hline
\textbf{Term} & \textbf{Audio Samples} \\ 
\hline\hline
\multirow{3}{4em}{4-fold}  & 2,528 (AP) \\
& 2,092 (Pre-trained FS2) \\
& 2,092 (Fine-tuned FS2) \\
\hline
16-fold & 8,368 (AP) \\
\hline
20-fold & 10,460 (AP + Fine-tuned FS2) \\
\hline
\end{tabular}
\end{table}

\subsection{Feature Extraction Method}
In our multimodal approach to sarcasm detection, we employ distinct feature extraction techniques for both textual and audio data. These techniques are pivotal in capturing the nuanced characteristics of sarcasm from each modality. Below, we detail the specific methods used for extracting features from text and audio inputs, highlighting their unique roles in our analysis.

\subsubsection{Textual Feature Extraction}
For encoding textual data, we utilize the Bidirectional Encoder Representations from Transformers (BERT)~\cite{bert}, pre-trained on the Toronto BookCorpus\footnote{https://en.wikipedia.org/wiki/BERT\_(language\_model)}(800M words) and English Wikipedia\footnote{https://en.wikipedia.org/wiki/English\_Wikipedia}(2,500M words). BERT is instrumental in our approach due to its proven capability in understanding and processing natural language, making it highly effective for generating context-aware embeddings for English text. This effectiveness in capturing textual nuances is critical, as demonstrated in various multimodal sarcasm detection studies~\cite{Castro, Pramanick, Ding}.

In the preprocessing phase, each text input $t_i=\{t_o,t_b^L\}$, 
is framed with $(CLS)$ and $(SEP)$ tokens to mark the start and end. This procedure is described as:
\begin{equation}
\label{eq3}
B=BERT\_Tokenizer([CLS]+t_i+[SEP])
\end{equation}
Here, $B$ denotes the token set derived from the input text sample. We standardize the token sequence length ($S$) to a maximum of 20, determined by calculating the mean of the second and third quartiles of the lengths of all text samples. The tokenized input is then processed through the BERT encoder:
\begin{equation}
\label{eq4}
T_f=BERT(B)
\end{equation}
Since the BERT model generates a 768-dimensional vector for each token, the resultant textual feature matrix is $T_f \in \mathbb{R}^{S\times D}$, where, $S=20$ is the number of tokens and $D=768$, represents the embedding dimension for every token. The textual feature matrix is later forwarded to the attention module for further processing. 

\subsubsection{Audio Feature Extraction}
Our audio feature extraction leverages VGGish~\cite{Hershey}, a Convolutional Neural Network (CNN)-based model specifically designed for audio tasks, pre-trained on the expansive YouTube-8M dataset\footnote{https://research.google.com/youtube8m/}. VGGish’s effectiveness in audio classification, particularly in sarcasm detection, has been validated in prior research~\cite{Gao, Liu, YZhang2, YZhang}. For each audio input $a_i=\{a_o,a_{AP}^L,a_{FS}^L\}$, we apply VGGish as follows:
\begin{equation}
\label{eq5}
A_f=VGGish(a_i)
\end{equation}
Each audio input $a_i$ undergoes resampling to 16 kHz, mono format, and conversion into a mel-spectrogram using a 25 ms window and 10 ms hop length. A 64 mel bin is applied to this spectrogram, followed by a $\log_2$ transformation. The data is divided into 0.96-second segments of 96 $\times$ 64 dimensions. These segments are processed through VGGish, where we utilize its pre-trained weights, excluding the last two fully-connected layers, to extract a feature matrix $A_f$ of dimensions 24 $\times$ 512. This matrix is then prepared for integration with the textual features.

\subsection{Feature Fusion and Classification}
Self-attention models are a class of neural networks that use attention mechanisms to discern relationships within a sequence. Unlike traditional models, they do not depend on recurrence or convolution. Their capability to capture long-range dependencies and parallelize computations enhances both accuracy and computational efficiency. However, these models also present challenges, such as increased memory and computational demands, sensitivity to positional encoding, and limited interpretability. The self-attention mechanism functions by generating a weight vector, which is then applied to the hidden embeddings within the sequence. This process prioritizes more significant elements in the sequence.

In our experimental approach, we first adjust the dimensions of the textual feature matrix $T_f$ from (20, 768) to (20, 512) to align with the dimensions of the acoustic features. 
This adjustment to match the dimensions of the acoustic features is essential, as maintaining a consistent feature vector length across different modalities is necessary before implementing the self-attention mechanism.
The self-attention mechanism is integrated into our model architecture to enable the model to selectively attend to different parts of the input sequence for each element in the output sequence. This enhances the model's ability to capture dependencies across varying distances in sequential data.
Let  $M$ be the modality-specific feature matrix where $M \in \mathbb{R}^{N\times 512}$, and 
$m$ is the modality indicator where $m \in \{a,t\}$. Here, $a$ and $t$ denote the acoustic and textual modality respectively.
\begin{equation}
\tilde{m}=Self-Attention(M)
\end{equation}
The detailed procedure of the self-attention mechanism is illustrated below. The input matrix $M$ is linearly transformed into three matrices $Q, K$, and $V$ using the learnable weight matrices $W_Q, W_K$, and $W_V$ respectively.

\begin{equation}
Q=MW_Q
\end{equation}
\begin{equation}
K=MW_K
\end{equation}
\begin{equation}
V=MW_V\\
\end{equation}
Then, attention scores are calculated by taking the dot product of the query $Q$ and key matrices $K$, scaled by $\sqrt{d_k}$, where $d_k$ is the dimension of the key vectors.
\begin{equation}
\text{Attention}(Q, K, V) = \text{softmax}\left(\frac{QK^T}{\sqrt{d_k}}\right) \cdot V
\end{equation}
The final output is obtained by applying the attention scores to the value matrix $V$ and linearly transforming the result using a weight matrix $W_O$.
\begin{equation}
\tilde{m}= Attention(Q,K,V)W_O
\end{equation}
Here, $W_O$ is another learnable matrix.
Finally, we use skip connections to obtain the final feature vector $(\tilde{m_s})$. Skip connections, used alongside attention mechanisms, ensure that while the network focuses on specific features, it also retains the broader context from the original input, leading to more balanced and effective learning.
\begin{equation}
\tilde{m_s}= \sum_{i=1}^{N} (M_{i} \cdot \tilde{m})
\end{equation}
We apply this mechanism individually on the textual and acoustic modality to obtain the self-attended textual feature vector ($\tilde{t_s}\in \mathbb{R}^{512}$) and self-attended acoustic feature vector ($\tilde{a_s}\in \mathbb{R}^{512}$) respectively.
The attended representations from both text and audio modalities are then concatenated as shown in Equation~\ref{eq:fusion}.
\begin{equation}
\tilde{p}=\tilde{t_s}\oplus\tilde{a_s}
\label{eq:fusion}
\end{equation}
Here, $\oplus$ signifies concatenation, producing an overall feature vector $\tilde{p}$ with a dimension of 1024. This combined feature vector is then fed into fully-connected layers. The training process involves updating gradients based on the binary cross-entropy loss function, and predictions about the sarcasm class are determined using a sigmoid operation:
\begin{equation}
y=\sigma({\tilde{p}})
\end{equation}
In this equation, $y$ represents the predicted class, and $\tilde{p}$ is the concatenated feature vector.

\section{Experiments}
In this section, we provide a detailed overview of our experimental implementation and present the results of the experiments. 

\subsection{Experimental Setup}
This subsection outlines our experimental framework, detailing the dataset utilized, comparison benchmarks, training specifics, and evaluation metrics. 

\subsubsection{Dataset Description}
Our experiments were conducted using the MUStARD dataset~\cite{Castro}, comprising 690 audiovisual utterances from American TV sitcoms, evenly split between sarcastic and non-sarcastic instances. This multimodal dataset is particularly suited for sarcasm detection as it includes contextual information and annotated sarcasm labels. Additionally, it addresses speaker bias by including samples from the same speakers across both sarcasm categories. To investigate the impact of varying data augmentation volumes, we created augmented datasets using bimodal data augmentation techniques and evaluated different audio synthesis methods. Table~\ref{table3} provides a comprehensive overview of the data used in our experiments.

\begin{table}[t]
\caption{Data Statistics\label{table3}}
\small
\centering
\begin{tabular}{|p{12em} | p{5em} | p{5em}|}
\hline
\textbf{Augmentation} & \textbf{Modality} & \textbf{Samples} \\ 
\hline\hline
\multirow{2}{12em}{No} & Text & 690 \\
   & Audio & 690 \\
\hline
\multirow{2}{12em}{4-fold(AP)} & Text & 3,218 \\
           & Audio & 3,218 \\
\hline
\multirow{2}{12em}{4-fold(Pre-trained FS2)} & Text & 2,782 \\
                        & Audio & 2,782 \\
\hline
\multirow{2}{12em}{4-fold(Fine-tuned FS2)}  & Text & 2,782 \\
                        & Audio & 2,782 \\
\hline                        
\multirow{2}{12em}{16-fold(AP)} & Text & 9,058 \\
            & Audio & 9,058 \\
\hline            
\multirow{2}{12em}{20-fold(AP+Fine-tuned FS2)} & Text & 11,150 \\
                           & Audio & 11,150 \\
\hline
\end{tabular}
\end{table}

\subsubsection{Benchmarked Methods}
The methods against which AMuSeD is benchmarked follow:

\textbf{SVM}~\cite{Cortes}: SVM is a widely recognized binary linear classifier in multimodal fusion research, known for its robustness and effectiveness in binary classification tasks.

\textbf{Inter-segment Inter-modal Attention and Intra-segment Inter-modal Attention (Ie Attention \& Ia Attention)}~\cite{Chauhan}: These dual attention mechanisms focus on capturing relationships both within (intra) and across (inter) different modalities. Their outputs are applied sequentially, concatenated, and then used as input for a softmax layer, which performs the final classification task.

\textbf{Contrastive-Attention-based Sarcasm Detection (ConAttSD)}~\cite{XZhang}: ConAttSD leverages a unique inter-modality contrastive attention mechanism. This method emphasizes the importance of incongruity between different modalities, which is a crucial aspect of sarcasm detection.

\textbf{M2Seq2Seq}~\cite{YZhang2}: M2Seq2Seq is designed with dual attention mechanisms within its encoder. It enables concurrent learning of both intra-modal and inter-modal dynamics, leading to multi-label classification that encompasses sentiment, emotion, and sarcasm.

\textbf{Gated Emoji-aware Multimodal Attention (GEMA)}~\cite{Chauhan2}: GEMA is an innovative deep learning framework that integrates emoji-awareness into multimodal multitasking. It assigns greater significance to features linked with emojis, thus enhancing the relevancy of input data for the softmax classification layer.

\subsubsection{Training Details}
We train our post self-attention features by a fully-connected neural network. Table~\ref{tabletrain} shows the hyper-parameters used in our training setup. 

\begin{table}[H]
\caption{Hyper-parameters in Training\label{tabletrain}}
\centering
\small
\begin{tabular}{|p{8em} | p{15em}|}
\hline
\textbf{Hyper-parameters} & \textbf{Value} \\ 
\hline\hline
Dense layer & 512, Dropout~\cite{Srivastava} = 0.5 \\
Activation   & ReLU~\cite{Nair} \\
Optimizer & Adam~\cite{Kingma} (lr = $10^{-4}$) \\
Output layer & Sigmoid \\
Loss function & Binary Cross-entropy \\
Batch & [16, 32, 64, 128, 256] \\
Epoch & 2000, Early stopping = 50 \\
\hline
\end{tabular}
\end{table}

\subsubsection{Evaluation Metrics}
The evaluation of our model employs standard metrics: precision (P), recall (R), and F1-score (F1), all of which hinge on the counts of true positives (TP), false positives (FP), true negatives (TN), and false negatives (FN). These metrics are widely used in sarcasm detection research and have been consistently applied in studies using the MUStARD dataset, facilitating a reliable comparison of our approach against existing methods~\cite{Castro, Wu, Pramanick}. In alignment with the evaluation protocol established by Castro~\textit{et al}.~\cite{Castro}, our experiment uses a 5-fold cross-validation method. The dataset is split according to a predefined file accessible on GitHub\footnote{https://github.com/soujanyaporia/MUStARD}. Our evaluation follows a speaker-dependent model, where the same speakers appear in both training and testing datasets, a common aspect in sarcasm detection research.

\subsection{Experimental Results}
This subsection provides a detailed analysis of the experiments conducted to evaluate the effectiveness of AMuSeD. We compare its performance against state-of-the-art methods, focusing on factors such as the volume of augmented data, the impact of different speech synthesizers, and the efficiency of diverse attention mechanisms.

\subsubsection{Overall Model Performance}
AMuSeD, which uses both text and audio modalities augmented through our bimodal data augmentation process, is evaluated against different benchmark methods. The comparative results are presented in Table~\ref{table4}. 
\begin{table}[h]
\centering
\small
\caption{Comparison Results against benchmarked Methods}
\label{table4}
\begin{tabular}{|p{12em} | p{3em} | p{3em} | p{3em} |}
\hline
\textbf{Model} & \textbf{P} & \textbf{R} & \textbf{F1} \\ 
\hline\hline
SVM & 72.0 & 71.6 & 71.6 \\
IeAttention\&Ia-Attention & 72.1 & 71.6 & 72.0 \\
ConAttSD & 72.5 & 74.0 & 74.0 \\
M2Seq2Seq & 74.5 & 74.7 & 74.6 \\
GEMA & 77.9 & 76.9 & 76.7 \\
Proposed Model & 81.6 & 81.0 & 81.0 \\
\hline
\end{tabular}
\end{table}
Notably, our approach achieves an F1-score of 81.0\%, a significant outcome even when compared to models that incorporate three modalities from the dataset. This comparison highlights the superior efficacy of attention-based models, including ours, over traditional  methods like SVM. This superiority is primarily attributed to the proficiency of attention mechanisms in capturing complex relationships within each modality. The following sections offer more detailed insights into how AMuSeD’s self-attention mechanism intricately processes feature vectors by considering intra-modal relationships, and how it stands in contrast to other models employing attention mechanisms.

\subsubsection{Data Augmentation Analysis}
\begin{figure}[t]
\centering
\includegraphics[width=0.45\textwidth]{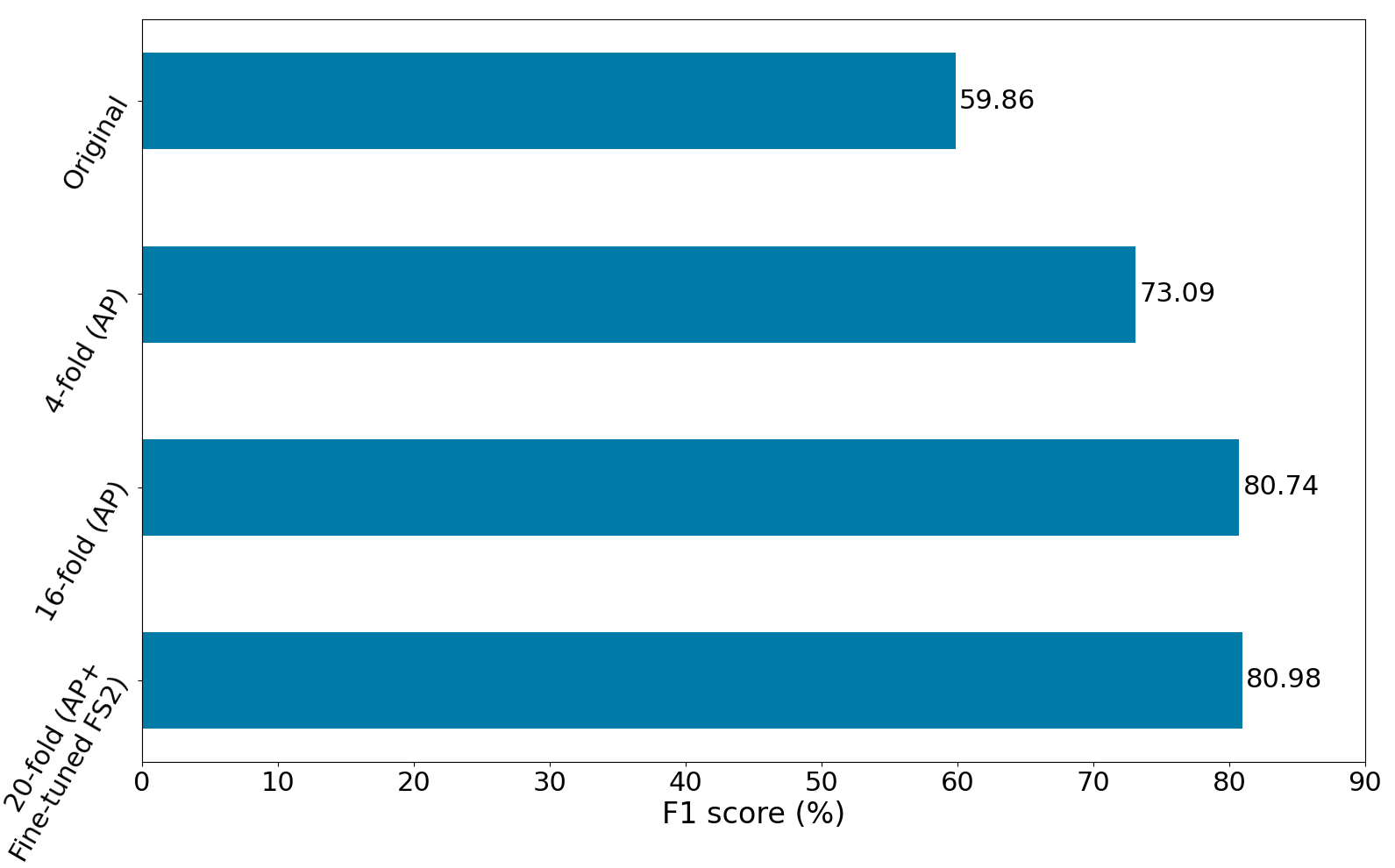}
\caption{The effect of augmented data size.}
\label{fig1}
\end{figure}

\begin{figure}[t]
\centering
\includegraphics[width=0.5\textwidth]{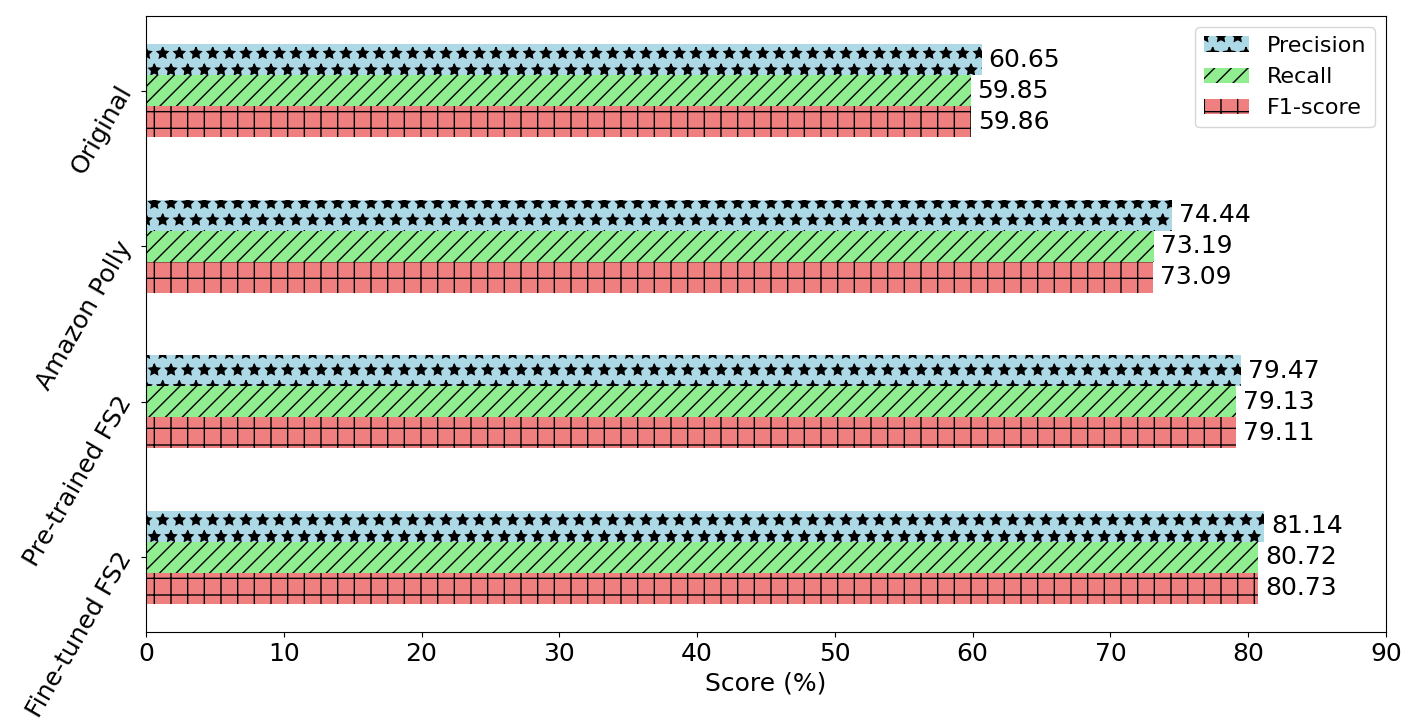}
\caption{The effect of synthesizers.}
\label{fig2}
\end{figure}

To explore the influence of data augmentation on model performance, we conducted experiments focusing on the size of the augmented dataset. The relationship between the volume of augmented data and model efficacy is depicted in Figure~\ref{fig1}. Our findings indicate a clear positive trend: as the volume of data increases, so does the performance of the model. Notably, when employing the same synthesizer, the model utilizing 16-fold augmented data demonstrated superior performance compared to the 4-fold augmentation. Furthermore, the model with 20-fold augmented data achieved the highest F1-score, reaching 80.98\%, which is approximated to 81.0\%. This result strongly supports the effectiveness of our data augmentation strategy in enhancing the model’s sarcasm detection capabilities.
Our investigation extended to the impact of different speech synthesizers on the 4-fold augmented data. Figure~\ref{fig2} presents the results of this analysis. We observed that the choice of synthesizer significantly influenced model performance. Specifically, the two FS2 based synthesis methods outperformed AP, with the Fine-tuned FS2 version achieving the highest F1-score. This suggests that FS2, especially when fine-tuned, is more effective in capturing speech prosody elements essential for sarcasm classification.

\begin{figure}[t]
\centering
\captionsetup[subfloat]{labelfont=scriptsize,textfont=scriptsize}
\subfloat[Quality.]{\includegraphics[width=3.5in]{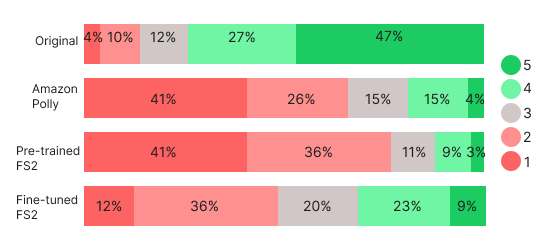}%
\label{MOS for quality}}
\hfil
\subfloat[Sarcastic similarity.]{\includegraphics[width=3.5in]{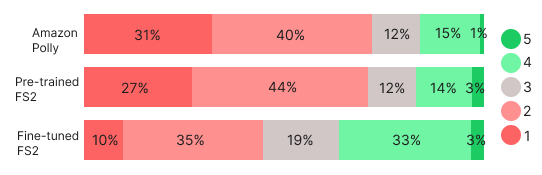}%
\label{MOS for sarcastic similarity}}
\caption{Subjective evaluation from the listening test. This figure indicates the percentage of users who provided ratings 1 (the lowest) to 5 (the highest) for each model.}
\label{mosresults}
\end{figure}

Furthermore, we conducted a subjective evaluation to assess the perceived quality and sarcasm resemblance of audio generated by different synthesizers. For this, we created a listening test comprising 10 natural speech utterances (original group) and their back-translated versions synthesized by AP, Pre-trained FS2, and Fine-tuned FS2 (contrastive group). Fifty listeners with no reported hearing impairments participated in this evaluation. They rated each utterance based on overall quality and sarcasm resemblance on a scale from 1 to 5, using the mean opinion score (MOS) method. The results, depicted in Figure~\ref{mosresults}, reveal that Fine-tuned FS2 notably enhanced both speech quality and sarcasm resemblance, significantly outperforming AP and Pre-trained FS2.

\subsubsection{Significance of Attention Mechanisms}
Our analysis extends to models incorporating various attention strategies, including Parallel CoAttention, Word-level attention, Cross attention, and Self-attention. These models were evaluated using the 20-fold augmented data set. 
\begin{table}[H]
\caption{Comparison results with and without attention mechanisms\label{table5}}
\centering
\small
\begin{tabular}{|p{4em} |m{9.5em} | l | l | l|}
\hline
\textbf{Attention} & \textbf{Model} & \textbf{P} & \textbf{R} & \textbf{F1} \\ 
\hline\hline
\multirow{2}{4em}{No} & SVM & 60.83 & 58.99 & 58.36 \\
& Fully-connected Neural Networks & 79.84 & 79.42 & 79.41 \\
\hline
\multirow{4}{4em}{Yes} & Parallel CoAttention & 77.94  & 77.68 & 77.65 \\
& Word level attention & 77.74 & 77.39 & 77.31 \\
& Self-attention & 80.27 & 79.71 & 79.64 \\
& Cross attention & 81.10 & 80.87 & 80.88 \\
\hline
\end{tabular}
\end{table}
The outcomes of this comparative study, highlighting the effectiveness of each attention mechanism within our model framework, are concisely summarized in Table~\ref{table5}.

A detailed analysis of models with and without attention mechanisms reveals a performance disparity. Models incorporating attention mechanisms consistently outperform those without, highlighting the critical role these mechanisms play in improving classification accuracy. Further, we delved into the influence of skip connections on different models. This additional layer of analysis yielded model-specific results, as detailed in Table~\ref{table6}. Notably, while Cross attention models experienced a 1.91\% decrease in F1-score upon integrating skip connections, Self-attention models saw an increase of 1.34\% in their F1-score, achieving the highest score in our set of experiments. 

\begin{table}[H]
\caption{Influence of skip connections on Cross Attention and Self-attention \label{table6}}
\centering
\small
\begin{tabular}{|l | m{8em} | l | l| l |}
\hline
\textbf{Attention} & \textbf{Skip-connection} & \textbf{P} & \textbf{R} & \textbf{F1} \\ 
\hline\hline
\multirow{2}{2em}{Cross attention} &  No & 81.10 & 80.87 & 80.88 \\
&  Yes & 79.22 & 78.98 & 78.97 \\
\hline
\multirow{2}{2em}{Self-attention} &  No & 80.27  & 79.71 & 79.64 \\
&  Yes & 81.57 & 81.01 & 80.98 \\
\hline
\end{tabular}
\end{table}

\subsubsection{BERT vs. GLOVE}
Experiments analyzing text feature extractors were conducted to compare the performance of BERT~\cite{bert} and GLoVE~\cite{Jeffrey} on a fully-connected neural network. These experiments were carried out using both original and augmented datasets. The results, as shown in Table~\ref{table7}, indicate that BERT~\cite{bert} consistently achieves a higher F1-score compared to GLoVE~\cite{Jeffrey}, regardless of the dataset type. Particularly noteworthy is the enhanced performance of BERT~\cite{bert} with augmented data, underscoring its superior capability in extracting text features that are critical for effective sarcasm detection in our model.

\begin{table}[H]
\caption{BERT Vs. GLOVE as text feature extractor\label{table7}}
\centering
\small
\begin{tabular}{|l | m{8em} | l | l | l|}
\hline
\textbf{Data} & \textbf{Feature-Extractor} & \textbf{P} & \textbf{R} & \textbf{F1} \\ 
\hline\hline
\multirow{2}{4.2em}{Original} & BERT & 68.95 & 68.12 & 68.10 \\
& GLOVE & 68.56 & 67.97 & 67.99 \\
\hline
\multirow{2}{4.2em}{Augmented} & BERT & 72.82  & 72.32 & 72.36 \\
& GLOVE & 70.11 & 69.57 & 69.59 \\
\hline
\end{tabular}
\end{table}

\subsubsection{The Role of Multimodality}
The impact of multimodality on model performance is significant, highlighting the advantages of integrating text and audio data. Table~\ref{table8} illustrates this effect on a fully-connected neural network; the fusion of both text and audio modalities results in an F1-score of 68.10\%. This outcome accentuates the synergistic benefits derived from the multimodal approach, effectively combining different types of data to enhance sarcasm detection capabilities.

\begin{table}[H]
\caption{The effect of multimodality\label{table8}}
\centering
\small
\begin{tabular}{|m{8em} | m{4em} | m{4em} | m{4em}|}
\hline
\textbf{Modality} & \textbf{P} & \textbf{R} & \textbf{F1} \\ 
\hline\hline
Audio & 69.11 & 67.97 & 67.86 \\
Text & 68.12 & 64.35 & 66.18 \\
Text+Audio & 68.95 & 68.12 & 68.10 \\
\hline
\end{tabular}
\end{table}

\section{Discussion}
This study evaluates the effectiveness of our multimodal sarcasm detection method, AMuSeD, which introduces a novel text-audio bimodal data augmentation technique. We utilize Back Translation for text and speech synthesis for audio to produce aligned augmentations for these modalities. Our model employs self-attention with skip connections for effective features fusion to recognize sarcasm. Experimental results demonstrate the method’s efficacy, achieving an F1-score of 80.98\%, surpassing state-of-the-art models across the MUStARD dataset’s three modalities.

\subsection{Data Augmentation Insights}
Our data augmentation method’s effectiveness is validated by examining influential factors such as data size and synthesizer construction. Data size emerged as a critical determinant; larger datasets generally improved model performance. We explored this by generating 4-fold and 16-fold augmented data using the AP synthesizer. The results confirmed the positive impact of data size on performance. Additionally, synthesizer construction played a significant role. Within the 4-fold augmented data, audio samples generated by Fine-tuned FS2, capturing the sarcasm-specific prosody, outperformed other methods. This was corroborated by our listening test, where Fine-tuned FS2 scored highest in sarcasm similarity. These findings align with existing research~\cite{Rosenberg, Latif} and highlight the importance of fidelity in augmented data for sarcasm detection.

\subsection{Attention Mechanism Efficacy}
The effectiveness of self-attention mechanisms, particularly with skip connections, is evident in our multimodal sarcasm detection model. Self-attention aids in emphasizing relevant features within each modality, while skip connections preserve essential information across network layers. Interestingly, the application of skip connections in Cross Attention mechanisms resulted in a performance drop, emphasizing the nuanced interplay between text and audio modalities in sarcasm detection. This finding indicates that skip connections are not universally beneficial, as they can introduce complexity and potential noise into the model. In scenarios without data augmentation, attention-equipped models underperformed traditional methods like SVM and fully-connected neural networks. This suggests that attention mechanisms may not always be superior in text-audio sarcasm detection, especially in smaller datasets where SVM can be more effective. The adaptability of fully-connected neural networks in handling diverse data types is also evident in these results.

\subsection{Modality and Feature Analysis}
Our findings re-affirm the importance of multimodality in sarcasm detection, resonating with prior research in the field~\cite{Chauhan, XZhang, YZhang}. Sarcasm often relies on contextual information from multiple sources, necessitating a multimodal approach for accurate detection. Additionally, our results reinforce BERT’s~\cite{bert} efficacy in extracting text features for sarcasm detection. BERT’s transformer-based architecture allows it to capture context-dependent phenomena like sarcasm effectively, unlike GloVe~\cite{Jeffrey}, which generates static word embeddings without contextual information. This distinction is crucial for tasks like sarcasm detection, where context plays a pivotal role.

\subsection{Limitations}
Our experimental findings shed light on several topics requiring further exploration:

\textbf{Accuracy of Back Translation:} In capturing sarcasm’s subtleties, it is not always consistent, especially across languages with diverse syntactic and semantic structures. While translating between closely related languages might result in repetitive samples, very different languages can introduce translation errors, impacting the quality of the augmented dataset.

\textbf{Refinement of Data Augmentation Techniques:} While synthesized audio provides a means to augment data, it falls short in replicating the rich variety of human speech characteristics, such as accents, dialects, and unique speech patterns. This limitation is particularly significant in sarcasm detection, where nuances in vocal expression play a critical role. The quality of audio samples produced through our bimodal data augmentation process should be further refined. Ensuring these samples accurately encapsulate nuanced linguistic characteristics intrinsic to sarcasm is essential for the model’s effectiveness.

\textbf{Synchronization of Text and Audio Modalities:} Accurate alignment is crucial in sarcasm detection, where timing and intonation are key to understanding the context. Ensuring that these modalities are effectively synchronized to reflect the intended sarcasm is an area that requires further research and development.

\textbf{Integration of Video Modality:} A limitation of our current approach is the exclusion of video modality. While our study omits augmenting video data due to its potential overlap with image modality and alignment challenges with text and audio, the inclusion of video is vital. As indicated by Attardo~\textit{et al}.~\cite{Attardo} and De Vries~\textit{et al}.~\cite{Vries}, sarcasm often incorporates unique facial expressions, head movements, and body language. Future work should aim to integrate video data effectively, balancing the reduction of redundancy with the preservation of key sarcasm-related features.

\textbf{Enhancement of Contextual Information:} Incorporating contextual elements such as speaker information and surrounding utterances could significantly enhance our model. These factors are critical in sarcasm detection and should be integrated into future model architectures to improve their predictive accuracy.

\textbf{Exploring Advanced Model Architectures:} We recommend investigating additional model structures, particularly those like CoAttention, which are known for efficiently managing inter-modal interactions. These sophisticated architectures might offer superior performance and deeper insights into the dynamics of modality relationships.

\textbf{Across Cultural and Linguistic Application:} Sarcasm is a complex linguistic phenomenon that varies across different cultures and languages. The current methodology and datasets may not adequately represent this diversity, thereby limiting the model’s applicability in a global context.

\textbf{Improving Source Data Quality:} The quality and diversity of the source data heavily influence the effectiveness of our approach. Any inherent biases or lack of representativeness in the original dataset can adversely affect both the back-translation and audio synthesis processes.

\textbf{Computational Resources:} The computational resources required for processing and analyzing multimodal data are considerable. This includes the need for substantial processing power and memory, which could be a constraint in resource-limited settings. Acknowledging and addressing these computational requirements is essential for the scalability and practical deployment of our methodology.

\section{Conclusion}
This study marks a significant advance in sarcasm detection by employing a multimodal approach, addressing the unique challenges of sarcasm’s pragmatic and context-dependent nature. We have explored sarcasm detection, emphasizing the necessity of multimodality in interpreting sarcastic expressions that often convey meanings contrary to their literal interpretations. Our research answered two RQs:

\textbf{RQ1:} Does our innovative text-audio data augmentation approach enhance the detection of sarcasm in a multimodal context, and which key factors primarily determine its success? Our augmentation techniques, utilizing Back Translation and audio synthesis, are effective in boosting multimodal sarcasm detection. We investigated critical factors like data size and the sarcasm resemblance in synthesized speech, providing valuable insights into sarcasm detection methodologies.

\textbf{RQ2:} How does incorporating self-attention mechanisms with our augmentation strategy refine the model’s performance, thus pushing the boundaries of multimodal sarcasm detection? The integration of self-attention mechanisms with augmented data has showcased their efficacy in enhancing feature fusion for multimodal sarcasm detection, adding sophistication and potential for improved performance.

Our findings suggest that refining the prosodic qualities in synthesized audio can further enhance our data augmentation approach. Future research should focus on TTS techniques capable of producing speech that more accurately reflects sarcasm’s subtleties. Additionally, there is a promising research avenue in exploring the integration of video data within our bimodal augmentation framework, offering a more comprehensive view of sarcasm by incorporating visual cues alongside text and audio.

In conclusion, our research introduces an innovative text-audio bimodal data augmentation method that significantly boosts sarcasm detection capabilities. We also highlight the effectiveness of self-attention mechanisms in a multimodal context. Our study transcends traditional sentiment analysis boundaries, venturing into the integration of speech data and synthesis tools. This interdisciplinary effort tackles the complexities of sarcasm detection and paves the way for more nuanced and context-sensitive multimodal approaches in the field.

\section*{Acknowledgments}
We extend our deepest thanks to Vaibhav Chandra, Samarth Meghani, and Devraj Raghuvanshi for their essential contributions that significantly influenced the successful completion of this research. Our sincere appreciation also goes to Frank J. Hopwood for his invaluable insights, which greatly enhanced the depth and overall quality of our study.
 
\bibliographystyle{IEEEtran}
\bibliography{references}

\begin{thebibliography}{10}
\providecommand{\url}[1]{#1}
\csname url@samestyle\endcsname
\providecommand{\newblock}{\relax}
\providecommand{\bibinfo}[2]{#2}
\providecommand{\BIBentrySTDinterwordspacing}{\spaceskip=0pt\relax}
\providecommand{\BIBentryALTinterwordstretchfactor}{4}
\providecommand{\BIBentryALTinterwordspacing}{\spaceskip=\fontdimen2\font plus
\BIBentryALTinterwordstretchfactor\fontdimen3\font minus \fontdimen4\font\relax}
\providecommand{\BIBforeignlanguage}[2]{{%
\expandafter\ifx\csname l@#1\endcsname\relax
\typeout{** WARNING: IEEEtran.bst: No hyphenation pattern has been}%
\typeout{** loaded for the language `#1'. Using the pattern for}%
\typeout{** the default language instead.}%
\else
\language=\csname l@#1\endcsname
\fi
#2}}
\providecommand{\BIBdecl}{\relax}
\BIBdecl

\bibitem{Lishapeng}
S.~Li, A.~Chen, Y.~Chen, and P.~Tang, ``The role of auditory and visual cues in the interpretation of mandarin ironic speech,'' \emph{Journal of Pragmatics}, vol. 201, pp. 3--14, Nov. 2022.

\bibitem{kreuz1}
R.~J. Kreuz, M.~A. Kassler, L.~Coppenrath, and B.~McLain~Allen, ``Tag questions and common ground effects in the perception of verbal irony,'' \emph{Journal of Pragmatics}, vol.~31, no.~12, pp. 1685--1700, Nov. 1999.

\bibitem{kreuz2}
R.~J. Kreuz and K.~E. Link, ``Asymmetries in the use of verbal irony,'' \emph{Journal of Language and Social Psychology}, vol.~21, no.~2, pp. 127--143, June 2002.

\bibitem{Wilson}
D.~Sperber and D.~Wilson, ``Irony and use-mention distinction,'' in \emph{Radical Pragmatics}, P.~Cole, Ed.\hskip 1em plus 0.5em minus 0.4em\relax New York, NY, USA: Academic Press, 1981, pp. 295--318.

\bibitem{Gibbs}
R.~W. Gibbs, ``On the psycholinguistics of sarcasm,'' \emph{Journal of Experimental Psychology: General}, vol. 115, no.~1, pp. 3--15, March 1986.

\bibitem{Kreuz}
R.~J. Kreuz and S.~Glucksberg, ``How to be sarcastic: The reminder theory of verbal irony,'' \emph{Journal of Experimental Psychology: General}, vol. 118, pp. 347--386, Dec. 1989.

\bibitem{Schifanella}
R.~Schifanella, P.~de~Juan, J.~Tetreault, and L.~Cao, ``Detecting sarcasm in multimodal social platforms,'' in \emph{Proc. 24th ACM Int. Conf. Multimedia}, Amsterdam, The Netherlands, Oct. 2016, pp. 1136--1145.

\bibitem{Castro}
S.~Castro, D.~Hazarika, V.~Pérez-Rosas, R.~Zimmermann, R.~Mihalcea, and S.~Poria, ``Towards multimodal sarcasm detection (an obviously perfect paper),'' in \emph{Proc. 57th Annu. Meeting Assoc. Comput. Linguistics}, Florence, Italy, July 2019, pp. 4619--4629.

\bibitem{Wu}
Y.~Wu, Y.~Zhao, X.~Lu, B.~Qin, Y.~Wu, J.~Sheng, and J.~Li, ``Modeling incongruity between modalities for multimodal sarcasm detection,'' \emph{IEEE MultiMedia}, vol.~28, no.~2, pp. 86--95, April-June 2021.

\bibitem{Pramanick}
S.~Pramanick, A.~Roy, and V.~M.~P. Johns, ``Multimodal learning using optimal transport for sarcasm and humor detection,'' in \emph{2022 IEEE/CVF Winter Conf. Appl. Comput. Vis. (WACV)}, Waikoloa, HI, USA, Jan. 2022, pp. 546--556.

\bibitem{Vaswani}
A.~Vaswani, N.~Shazeer, N.~Parmar, J.~Uszkoreit, L.~Jones, A.~N. Gomez, L.~Kaiser, and I.~Polosukhin, ``Attention is all you need,'' in \emph{Proc. 31st Int. Conf. Neural Inf. Process. Syst.}, Long Beach, California, USA, June 2017, p. 6000–6010.

\bibitem{Rico}
R.~Sennrich, B.~Haddow, and A.~Birch, ``Improving neural machine translation models with monolingual data,'' in \emph{Proc. 54th Annu. Meeting Assoc. Comput. Linguistics (Volume 1: Long Papers)}, Berlin, Germany, Aug. 2016, pp. 86--96.

\bibitem{Marivate}
V.~Marivate and T.~Sefara, ``Improving short text classification through global augmentation methods,'' in \emph{Proc. Int. Cross-Domain Conf. Mach. Learn. and Knowl. Extraction (CD-MAKE)}, Canterbury, United Kingdom, Aug. 2020, pp. 385--399.

\bibitem{Ko}
T.~Ko, V.~Peddinti, D.~Povey, and S.~Khudanpur, ``Audio augmentation for speech recognition,'' in \emph{Interspeech 2015}, Dresden, Germany, Sep. 2015, pp. 3586--3589.

\bibitem{Li}
J.~Li, R.~T. Gadde, B.~Ginsburg, and V.~Lavrukhin, ``Training neural speech recognition systems with synthetic speech augmentation,'' \emph{ArXiv}, vol. abs/1811.00707, Nov. 2018.

\bibitem{Xu}
N.~Xu, W.~Mao, P.~Wei, and D.~Zeng, ``Mda: Multimodal data augmentation framework for boosting performance on sentiment/emotion classification tasks,'' \emph{IEEE Intelligent Systems}, vol.~36, no.~6, pp. 3--12, Nov.-Dec. 2021.

\bibitem{Ruixue}
R.~Tang, C.~Ma, W.~Zhang, Q.~Wu, and X.~Yang, ``Semantic equivalent adversarial data augmentation for visual question answering,'' in \emph{Proc. 16th Eur. Conf. Comput. Vis. (ECCV)}, Glasgow, United Kingdom, Nov. 2020, pp. 437--453.

\bibitem{Kumar}
S.~Kumar, A.~Kulkarni, M.~S. Akhtar, and T.~Chakraborty, ``When did you become so smart, oh wise one?! sarcasm explanation in multi-modal multi-party dialogues,'' in \emph{Proc. 60th Annu. Meeting Assoc. Comput. Linguistics}, Dublin, Ireland, May 2022, pp. 5956--5968.

\bibitem{Ray}
A.~Ray, S.~Mishra, A.~Nunna, and P.~Bhattacharyya, ``A multimodal corpus for emotion recognition in sarcasm,'' in \emph{Proc. 30th Lang. Resour. and Eval. Conf.}, Marseille, France, June 2022, pp. 6992--7003.

\bibitem{Ding}
N.~Ding, S.-w. Tian, and L.~Yu, ``A multimodal fusion method for sarcasm detection based on late fusion,'' \emph{Multimedia Tools and Applications}, vol.~81, pp. 8597--8616, Feb. 2022.

\bibitem{Hiremath}
B.~N. Hiremath and M.~M. Patil, ``Sarcasm detection using cognitive features of visual data by learning model,'' \emph{Expert Systems with Applications}, vol. 184, p. 115476, Dec. 2021.

\bibitem{Chauhan}
D.~S. Chauhan, D.~S. R, A.~Ekbal, and P.~Bhattacharyya, ``Sentiment and emotion help sarcasm? a multi-task learning framework for multi-modal sarcasm, sentiment and emotion analysis,'' in \emph{Proc. 58th Annu. Meeting Assoc. Comput. Linguistics}, Online, July 2020, pp. 4351--4360.

\bibitem{XZhang}
X.~Zhang, Y.~Chen, and G.~Li, ``Multi-modal sarcasm detection based on contrastive attention mechanism,'' in \emph{Proc. 10th Natural Lang. Process. and Chin. Comput.}, Qingdao, China, Oct 2021, pp. 822--833.

\bibitem{YZhang}
Y.~Zhang, Y.~Yu, D.~Zhao, Z.~Li, B.~Wang, Y.~Hou, P.~Tiwari, and J.~Qin, ``Learning multi-task commonness and uniqueness for multi-modal sarcasm detection and sentiment analysis in conversation,'' \emph{IEEE Transactions on Artificial Intelligence}, pp. 1--13, July 2023.

\bibitem{Aroyehun}
S.~T. Aroyehun and A.~Gelbukh, ``Aggression detection in social media: Using deep neural networks, data augmentation, and pseudo labeling,'' in \emph{Proc. 1st Workshop Trolling, Aggression and Cyberbullying ({TRAC})}, Santa Fe, New Mexico, USA, Aug. 2018, pp. 90--97.

\bibitem{Lee}
H.~Lee, Y.~Yu, and G.~Kim, ``Augmenting data for sarcasm detection with unlabeled conversation context,'' in \emph{Proc. 2nd Workshop Figurative Lang. Process.}, July 2020, pp. 12--17.

\bibitem{Gao}
X.~Gao, S.~Nayak, and M.~Coler, ``Deep cnn-based inductive transfer learning for sarcasm detection in speech,'' in \emph{Interspeech 2022}, Incheon, Republic of Korea, Sep. 2022, pp. 2323--2327.

\bibitem{LibriTTS}
H.~Zen, V.-T. Dang, R.~A.~J. Clark, Y.~Zhang, R.~J. Weiss, Y.~Jia, Z.~Chen, and Y.~Wu, ``Libritts: A corpus derived from librispeech for text-to-speech,'' in \emph{Interspeech 2019}, Graz, Austria, Sep. 2019.

\bibitem{Rosenberg}
A.~Rosenberg, Y.~Zhang, B.~Ramabhadran, Y.~Jia, P.~Moreno, Y.~Wu, and Z.~Wu, ``Speech recognition with augmented synthesized speech,'' in \emph{Proc. IEEE Automatic Speech Recognit. and Understanding Workshop (ASRU)}, Sentosa, Singapore, Dec. 2019, pp. 996--1002.

\bibitem{Huang}
J.~Huang, Y.~Li, J.~Tao, Z.~Lian, M.~Niu, and M.~Yang, ``Multimodal continuous emotion recognition with data augmentation using recurrent neural networks,'' in \emph{Proc. Audio/Vis. Emotion Challenge and Workshop}, Seoul, Republic of Korea, Oct. 2018, pp. 57--64.

\bibitem{Hao}
X.~Hao, Y.~Zhu, S.~Appalaraju, A.~Zhang, W.~Zhang, B.~Li, and M.~Li, ``Mixgen: A new multi-modal data augmentation,'' in \emph{Proc. IEEE/CVF Winter Conf. Appl. Comput. Vis. Workshops (WACVW)}, Waikoloa, Hawaii, Feb. 2023, pp. 379--389.

\bibitem{bert}
J.~Devlin, M.-W. Chang, K.~Lee, and K.~Toutanova, ``Bert: Pre-training of deep bidirectional transformers for language understanding,'' in \emph{Proc. North American Chapter Assoc. Comput. Linguistics (NAACL)}, Hyatt Regency, Minneapolis, USA, June 2019, pp. 4171--4186.

\bibitem{Hershey}
S.~Hershey and et~al, ``Cnn architectures for large-scale audio classification,'' in \emph{Proc. 42nd IEEE Int. Conf. Acoust., Speech and Signal Process. (ICASSP)}, New Orleans, USA, March 2017, pp. 131--135.

\bibitem{FastSpeech}
C.-M. Chien, J.-H. Lin, C.-y. Huang, P.-c. Hsu, and H.-y. Lee, ``Investigating on incorporating pretrained and learnable speaker representations for multi-speaker multi-style text-to-speech,'' in \emph{Proc. IEEE Int. Conf. Acoust., Speech and Signal Process. (ICASSP)}, Toronto, Ontario, Canada, March 2021, pp. 8588--8592.

\bibitem{Kingma}
D.~P. Kingma and J.~Ba, ``Adam: A method for stochastic optimization,'' in \emph{Proc. 3rd Int. Conf. Learn. Representations (ICLR)}, San Diego, CA, USA, Dec. 2011.

\bibitem{HiFi}
J.~Kong, J.~Kim, and J.~Bae, ``Hifi-gan: Generative adversarial networks for efficient and high fidelity speech synthesis,'' in \emph{Proc. 34th Int. Conf. Neural Inf. Process. Systems}, Vancouver, BC, Canada, Oct. 2020.

\bibitem{Liu}
Y.~Liu, Y.~Zhang, and D.~Song, ``A quantum probability driven framework for joint multi-modal sarcasm, sentiment and emotion analysis,'' \emph{IEEE Transactions on Affective Computing}, pp. 1--15, 2023.

\bibitem{YZhang2}
Y.~Zhang, J.~Wang, Y.~Liu, L.~Rong, Q.~Zheng, D.~Song, P.~Tiwari, and J.~Qin, ``A multitask learning model for multimodal sarcasm, sentiment and emotion recognition in conversations,'' \emph{Information Fusion}, vol.~93, pp. 282--301, May 2023.

\bibitem{Cortes}
C.~Cortes and V.~Vapnik, ``Support-vector networks,'' \emph{Machine Learning}, vol.~20, no.~3, pp. 273--297, Sep. 1995.

\bibitem{Chauhan2}
D.~S. Chauhan, G.~V. Singh, A.~Arora, A.~Ekbal, and P.~Bhattacharyya, ``An emoji-aware multitask framework for multimodal sarcasm detection,'' \emph{Knowledge-Based Systems}, vol. 257, Dec. 2022.

\bibitem{Srivastava}
N.~Srivastava, G.~Hinton, A.~Krizhevsky, I.~Sutskever, and R.~Salakhutdinov, ``Dropout: A simple way to prevent neural networks from overfitting,'' \emph{Journal of Machine Learning Research}, vol.~15, no.~1, pp. 1929--1958, Jan. 2014.

\bibitem{Nair}
V.~Nair and G.~E. Hinton, ``Rectified linear units improve restricted boltzmann machines,'' in \emph{Proc. 27th Int. Conf. Mach. Learn.}, Haifa, Israel, June 2010, pp. 807--814.

\bibitem{Jeffrey}
J.~Pennington, R.~Socher, and C.~Manning, ``Glove: Global vectors for word representation,'' in \emph{Proc. 2014 Conf. Empirical Methods Natural Lang. Process. (EMNLP)}, Doha, Qatar, Oct 2014, pp. 1532--1543.

\bibitem{Latif}
S.~Latif, A.~Shahid, and J.~Qadir, ``Generative emotional ai for speech emotion recognition: The case for synthetic emotional speech augmentation,'' \emph{Applied Acoustics}, vol. 210, July 2023.

\bibitem{Attardo}
S.~Attardo, J.~Eisterhold, J.~Hay, and I.~Poggi, ``Multimodal markers of irony and sarcasm,'' \emph{Humor - International Journal of Humor Research}, vol.~16, no.~2, Jan. 2003.

\bibitem{Vries}
C.~de~Vries, B.~Oben, and G.~Brône, ``Exploring the role body in communicating ironic stance,'' \emph{Languages and Modalities}, vol.~1, pp. 65--80, Oct. 2021.

\end{thebibliography}

\begin{IEEEbiography}[{\includegraphics[width=1in,height=1.25in,clip,keepaspectratio]{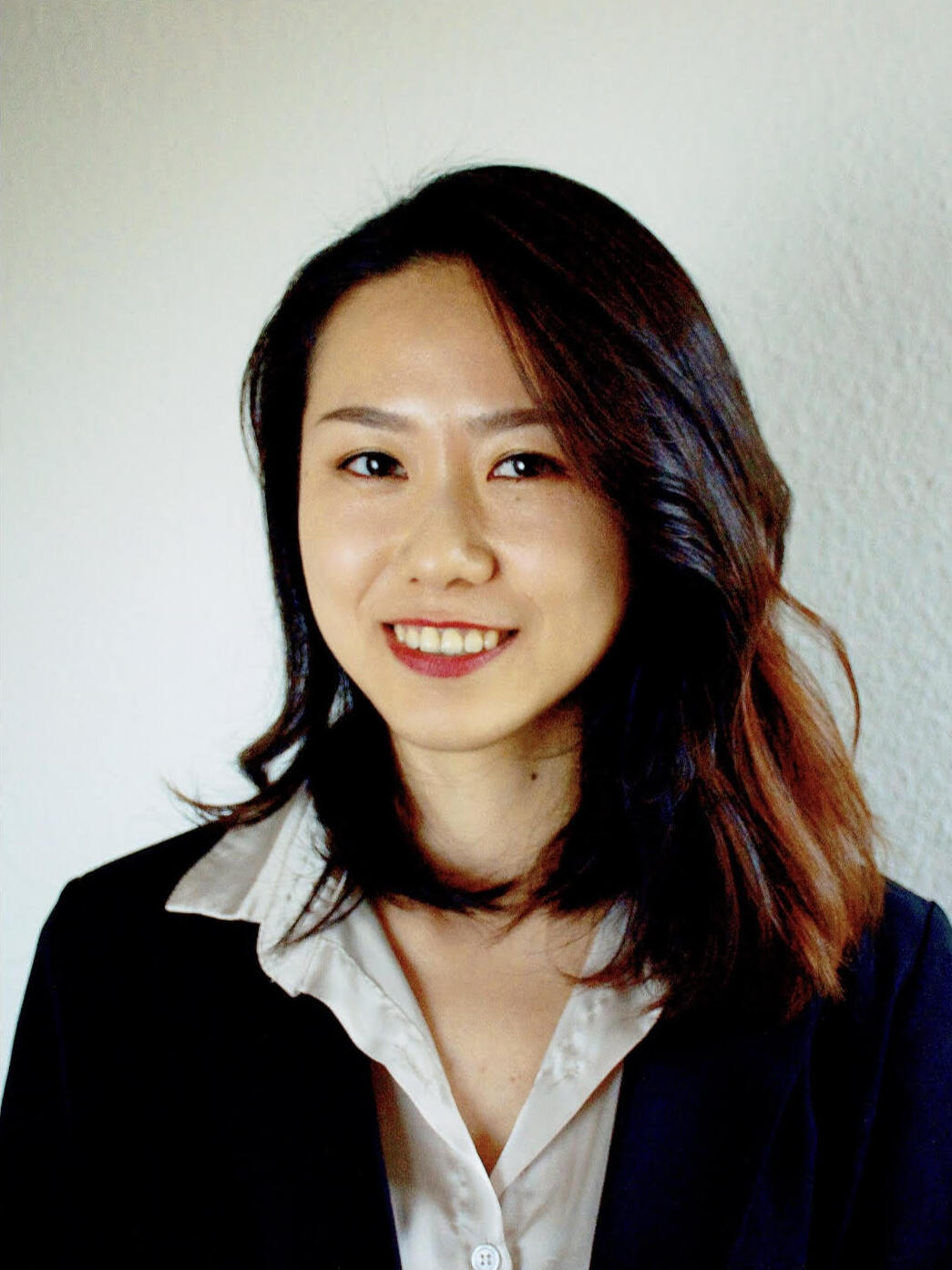}}]
{Xiyuan Gao} received the Master's degree in Speech and Language Processing from Konstanz University, Konstanz, Germany. She is currently a Ph.D. candidate from Speech Technology at the Faculty Campus Fryslân, University of Groningen, the Netherlands. Her research interests include linguistic insights driven sarcasm detection, multi-modal framework, speech technology.
\end{IEEEbiography}

\begin{IEEEbiography}[
{\includegraphics[width=1in,height=1.25in,clip,keepaspectratio]{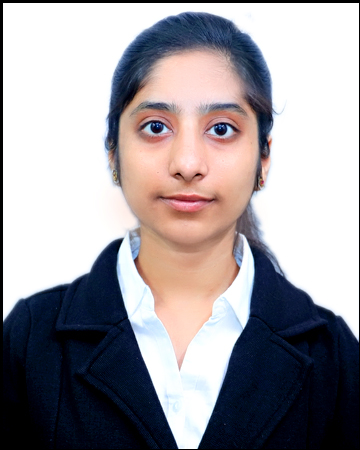}}
]
{Shubhi Bansal} received the B.Tech. degree in Computer Science and Engineering from Guru Gobind Singh Indraprastha University (GGSIPU), Delhi, India. She is currently pursuing the Ph.D. degree in the Department of Computer Science and Engineering at Indian Institute of Technology (IIT) Indore, Indore, India. Her research interests include natural language processing, social network analysis, deep learning, and artificial intelligence.
\end{IEEEbiography}

\begin{IEEEbiography}[
{\includegraphics[width=1in,height=1.25in,clip,keepaspectratio]{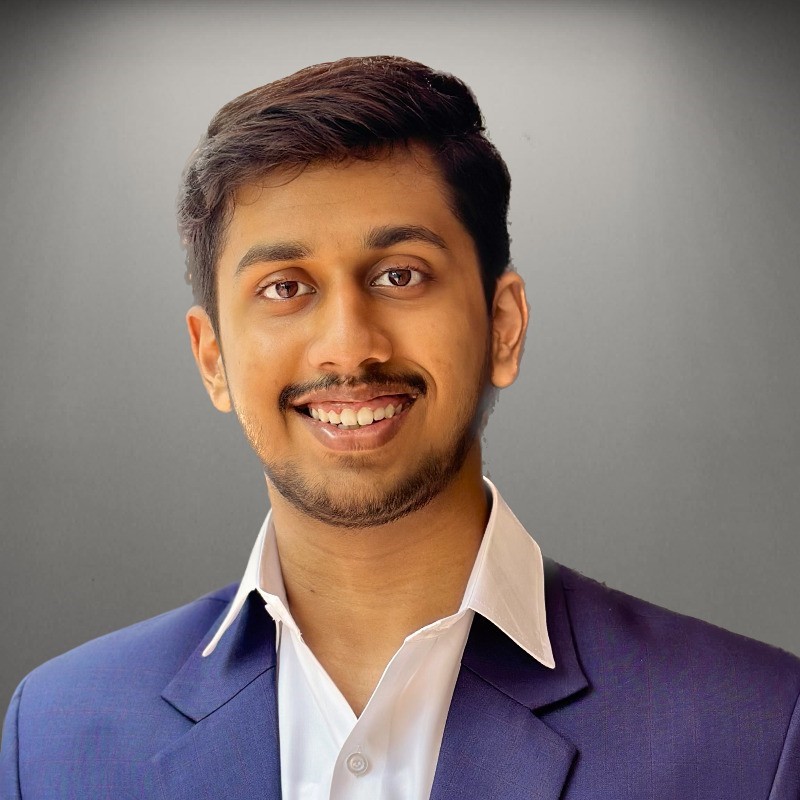}}
]
{Kushaan Gowda} is currently pursuing the M.S. degree in Computer Science at Columbia University, New York, United States. He received the B.Tech. in Computer Science and Engineering from Indian Institute of Technology (IIT) Indore, India, in 2023. His research interests include social networks, natural language processing, computer vision, and deep learning.

\end{IEEEbiography}

\begin{IEEEbiography}[
{\includegraphics[width=1in,height=1.25in,clip,keepaspectratio]{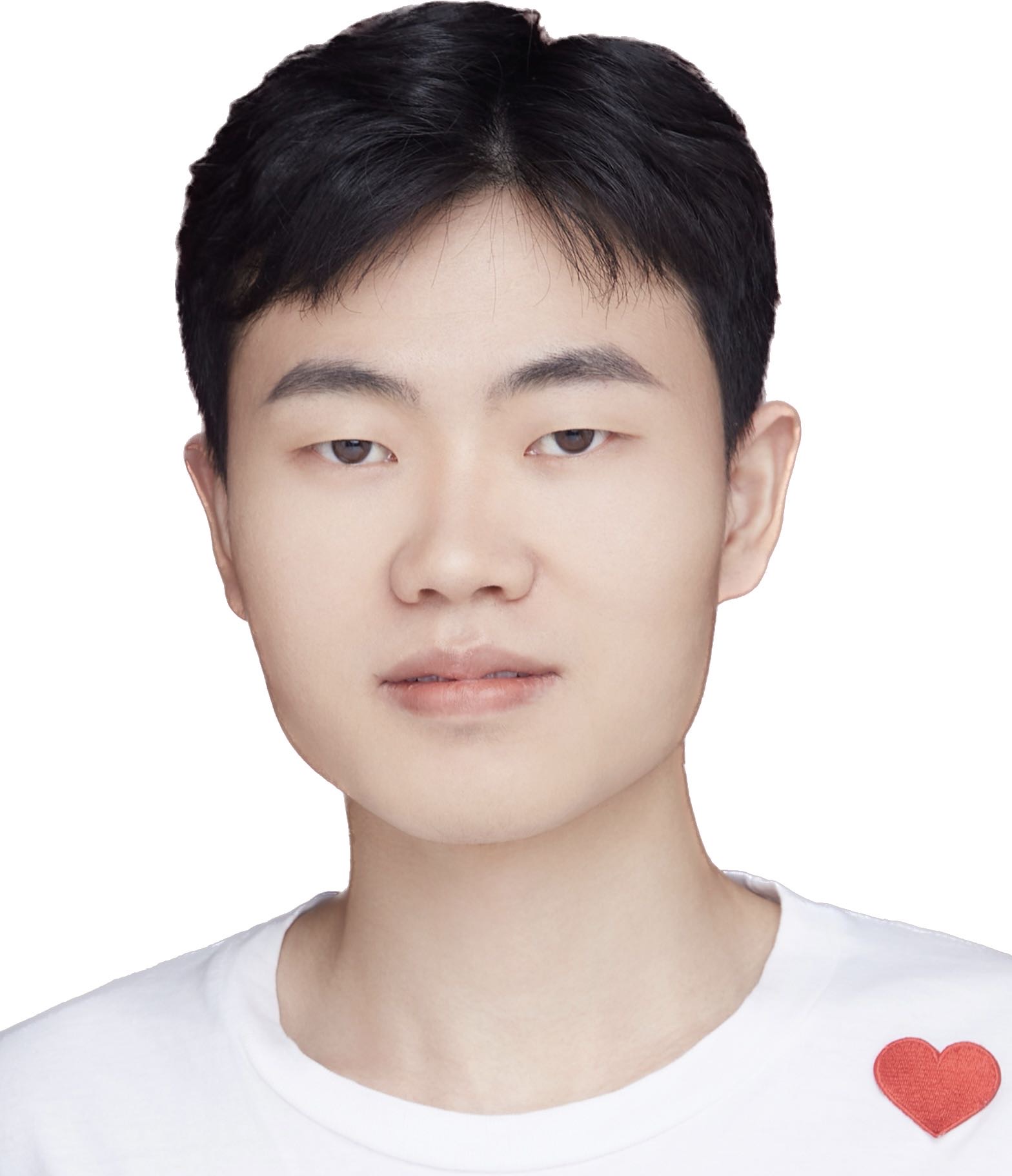}}
]
{Zhu Li} received the Master's degree in Software Engineering from Beijing Language and Culture University, Beijing, China. He is currently a Ph.D. candidate in Speech Technology at the Faulty Campus Fryslân, University of Groningen, the Netherlands. His research interests include expressive text-to-speech, voice conversion, large language models, and multi-modal framework.
\end{IEEEbiography}

\begin{IEEEbiography}[
{\includegraphics[width=1in,height=1.25in,clip]{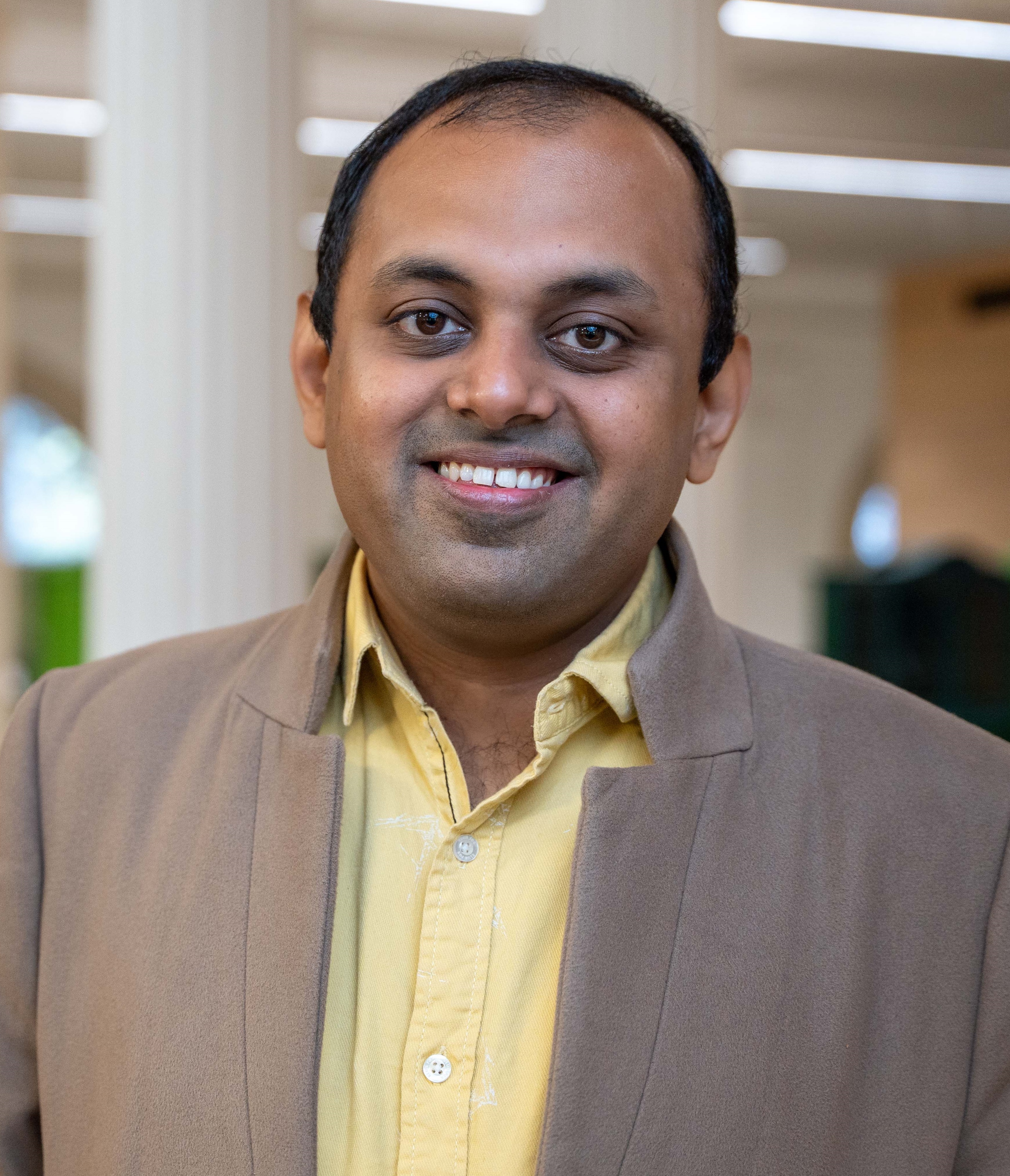}}
]
{Shekhar Nayak} received the Ph.D. degree and the M.Tech. degree from the Indian Institute of Technology (IIT) Hyderabad and IIT Delhi, India, in 2019 and 2011, respectively. He served as a Technology Consultant from 2011-2013 at Hewlett Packard Enterprise and as a Senior Chief Engineer at Samsung R\&D Institute, Bangalore, India from 2019-2021 in the Voice Services Department. He was a Research Assistant at the Institute for Infocomm Research (I2R), Agency for Science, Technology and Research (A*STAR), Singapore in 2015. Dr. Nayak is currently an Assistant Professor of Speech Technology at the Faculty Campus Fryslân, University of Groningen, the Netherlands since 2021. His research interests include speech recognition, speech synthesis and multi-modal signal processing and deep learning.
\end{IEEEbiography}

\begin{IEEEbiography}[
{\includegraphics[width=1in,height=1.25in,clip,keepaspectratio]{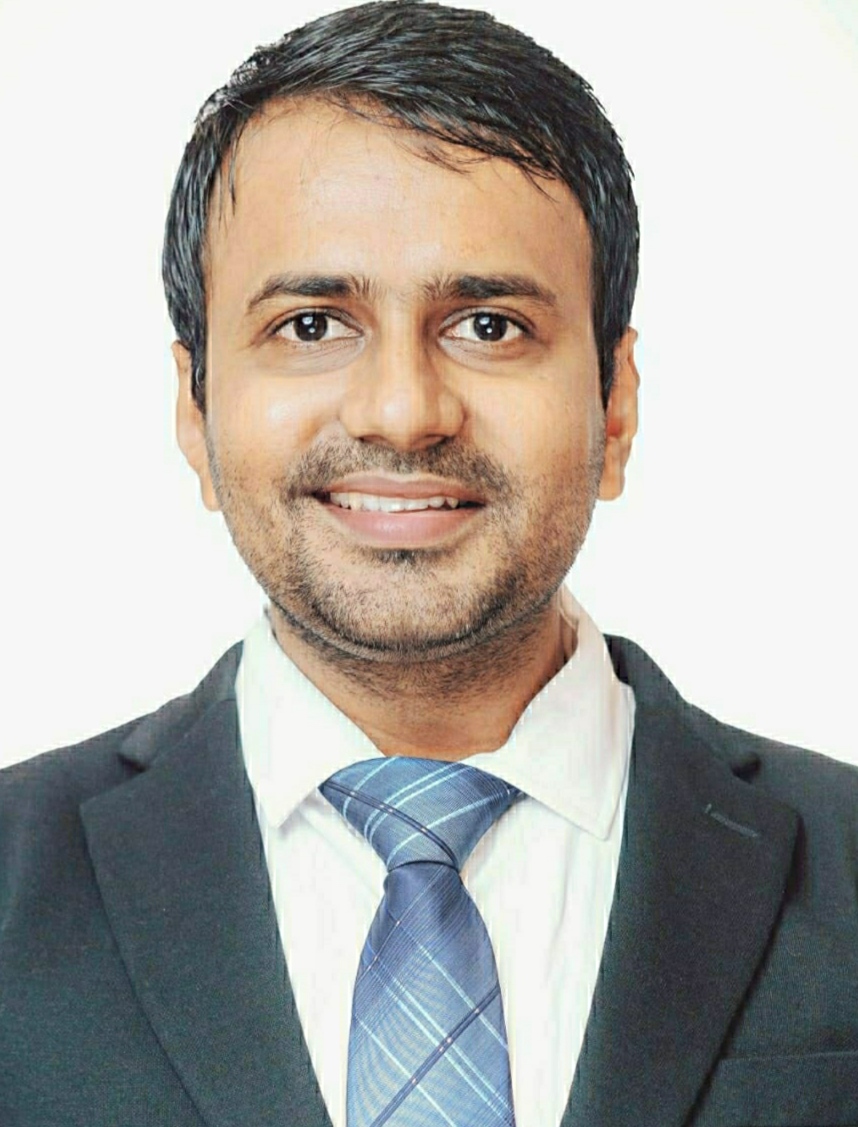}}
]
{Nagendra Kumar} received the Ph.D. degree from the Indian Institute of Technology (IIT) Hyderabad, Hyderabad, India, in 2019. He was a Scientist with the Institute for Infocomm Research (I2R), Agency for Science, Technology and Research (A*STAR), Singapore. He is currently an Assistant Professor at the Department of Computer Science and Engineering, IIT Indore, Indore, India. His research interests include natural language processing, social network analysis, deep learning, artificial intelligence, machine learning, and data mining.
\end{IEEEbiography}

\begin{IEEEbiography}[
{\includegraphics[width=1in,height=1.25in,clip,keepaspectratio]{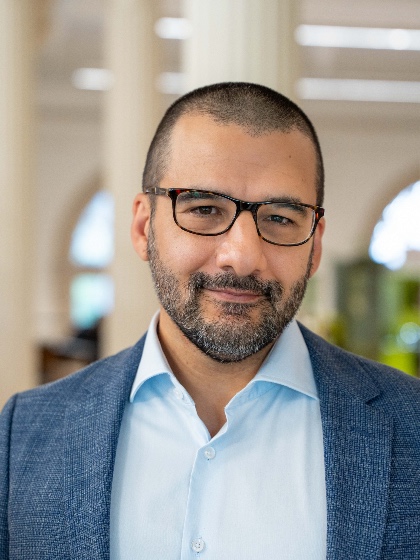}}
]
{Matt Coler} received the Ph.D. degree from the Free University of Amsterdam, the Netherlands, in 2010. After a postdoctoral appointment at his alma mater, he joined an AI start-up focusing on acoustic sensors, where he served as Head of the Cognitive Systems Unit. In 2012, he returned to academia. Dr. Coler is currently an Associate Professor of Speech Technology at the Faculty Campus Fryslân, University of Groningen, the Netherlands, where he works as the Director of the M.Sc. Voice Technology program and the Chair of the faculty Research Institute. His research interests include topics around speech technology with small data, low-resource languages, pragmatics in speech technology, and ethics. 
\end{IEEEbiography}

\end{document}